%% file: main.tex
\documentclass{article} 
\usepackage{iclr2026_conference,times}

\input{math_commands.tex}

\usepackage{amsthm}
\newtheorem{lemma}{Lemma}
\usepackage{hyperref}
\usepackage{url}

\usepackage[utf8]{inputenc} 
\usepackage[T1]{fontenc}    
\usepackage{hyperref}       
\usepackage{url}            
\usepackage{booktabs}       
\usepackage{amsfonts}       
\usepackage{nicefrac}       
\usepackage{microtype}      
\usepackage{xcolor}         
\usepackage{graphicx}
\usepackage{amssymb}
\usepackage{amsmath}
\usepackage{caption}
\usepackage{subcaption}
\usepackage[dvipsnames]{xcolor}
\usepackage{nicematrix}
\usepackage{wrapfig}
\usepackage{soul}
\usepackage{multirow} 
\usepackage[dvipsnames,svgnames,x11names,table]{xcolor}

\def\eg{\emph{e.g.}} 
\def\ie{\emph{i.e}} 
\newcommand{\Tref}[1]{Table~\ref{#1}}
\newcommand{\Eref}[1]{Eq.~(\ref{#1})}
\newcommand{\Fref}[1]{Fig.~\ref{#1}}
\newcommand{\Sref}[1]{Sec.~\ref{#1}}

\newcommand{\ours}{{RoSE}}
\newcommand{\dataset}{{MultiShade}}

\newcommand{\yingchen}[1]{{\color{red}{\bf\[yingchen]}}}

\newcommand{\blue}[1]{#1}

\title{Monocular Normal Estimation via Shading Sequence Estimation}

\author{
\begin{tabular}{c}
\textbf{Zongrui Li}\textsuperscript{1}\thanks{Equal contribution.} \quad
\textbf{Xinhua Ma}\textsuperscript{1}\footnotemark[1] \quad
\textbf{Minghui Hu}\textsuperscript{1} \quad
\textbf{Yunqing Zhao}\textsuperscript{2} \quad
\textbf{Yingchen Yu}\textsuperscript{2} \\
\textbf{Qian Zheng}\textsuperscript{3,4} \quad
\textbf{Chang Liu}\textsuperscript{5}\thanks{Corresponding author.} \quad
\textbf{Xudong Jiang}\textsuperscript{1} \quad
\textbf{Song Bai}\textsuperscript{2} \\
\mdseries
\textsuperscript{1}School of Electrical and Electronic Engineering, Nanyang Technological University \\
\mdseries
\textsuperscript{2}ByteDance \quad
\textsuperscript{3}College of Computer Science and Technology, Zhejiang University \\
\mdseries
\textsuperscript{4}The State Key Lab of Brain-Machine Intelligence, Zhejiang University \\
\mdseries
\textsuperscript{5}MoE Key Laboratory of Interdisciplinary Research of Computation and Economics, \\
\mdseries
Shanghai University of Finance and Economics \\
\mdseries
{\small \protect\url{https://github.com/LMozart/ICLR2026-RoSE.git}}
\end{tabular}
}

\iclrfinalcopy 

\begin{document}

\maketitle
\begin{center}
    \centering
    \captionsetup{type=figure}
    \includegraphics[width=1\textwidth]{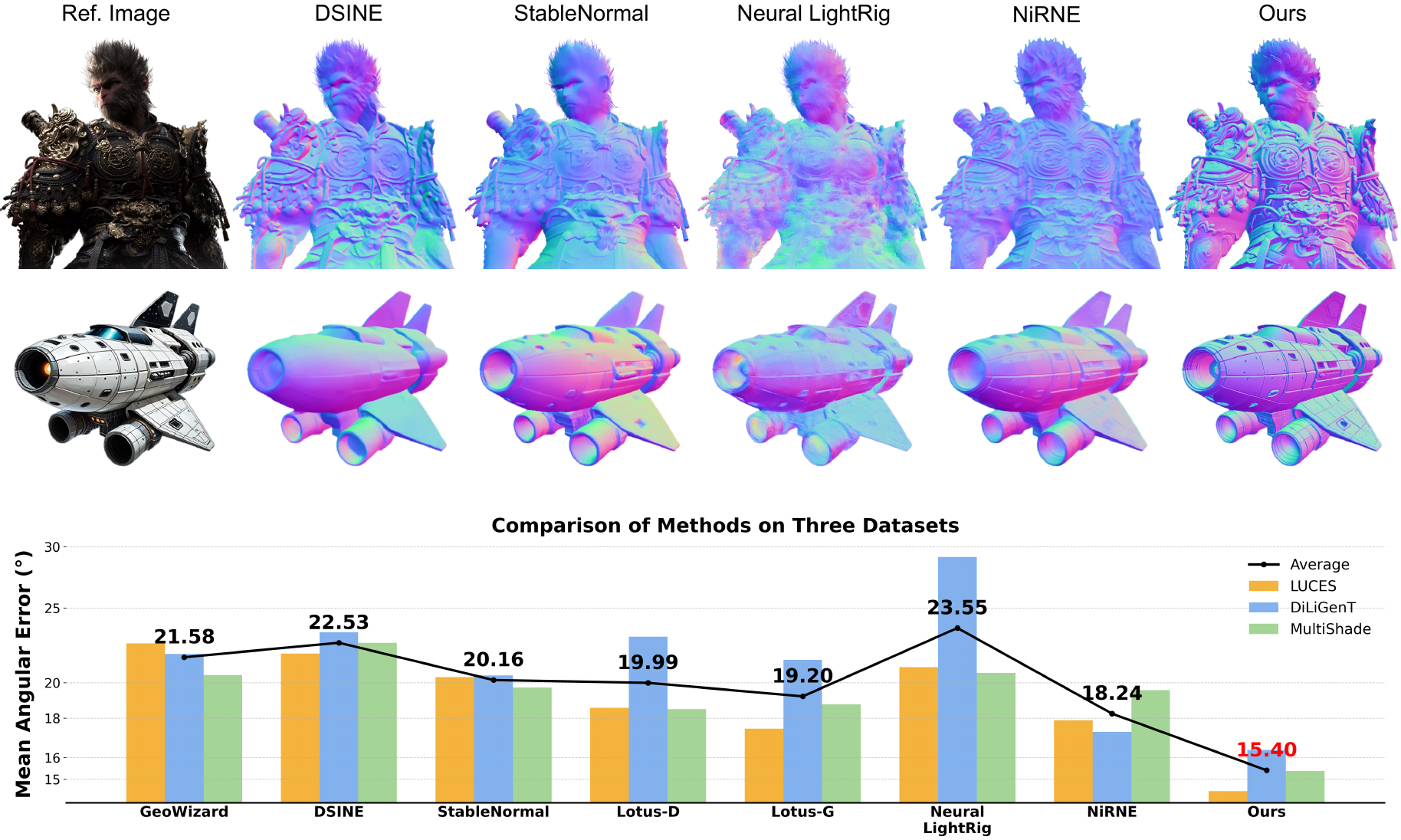}
    \captionof{figure}{We present \textbf{\ours}, a method using a video generative model for monocular normal map estimation, built on a new paradigm that reformulates normal estimation as a shading sequence estimation task. Results on complex and diverse scenarios show that \ours~reconstructs fine-grained geometric details and generalizes robustly to unseen datasets, achieving state-of-the-art performance in object-based monocular normal estimation on benchmark datasets.
}
    \label{fig:teaser}
\end{center}%

\begin{abstract}
    Monocular normal estimation aims to estimate the normal map from a single RGB image of an object under arbitrary lights. Existing methods rely on deep models to directly predict normal maps. However, they often suffer from \textit{3D misalignment}: while the estimated normal maps may appear to have a correct appearance, the reconstructed surfaces often fail to align with the 3D geometry. We argue that this misalignment stems from the current paradigm: the model struggles to distinguish and estimate varying geometry represented in normal maps, as the differences in underlying geometry are reflected only through relatively subtle color variations.
    To address this issue, we propose a new paradigm that reformulates normal estimation as shading sequence estimation, where shading sequences are more sensitive to various geometry information. By learning to infer the shading sequence of an object, the model can better capture underlying 3D geometry and thereby produce more accurate normal predictions. Building on this paradigm, we present \textbf{RoSE}, a method that leverages image-to-video generative models to predict shading sequences, which are then converted into normal maps by solving a simple ordinary least-squares problem. To enhance robustness and better handle complex objects, RoSE is trained on a synthetic dataset,~\dataset, with diverse shapes, materials, and light conditions. Experiments demonstrate that RoSE achieves state-of-the-art performance on both synthetic and real-world benchmark datasets for object-based monocular normal estimation. 
    
\end{abstract} \vspace{-0.12in}

\begin{figure}
    \centering
    \includegraphics[width=0.9\linewidth]{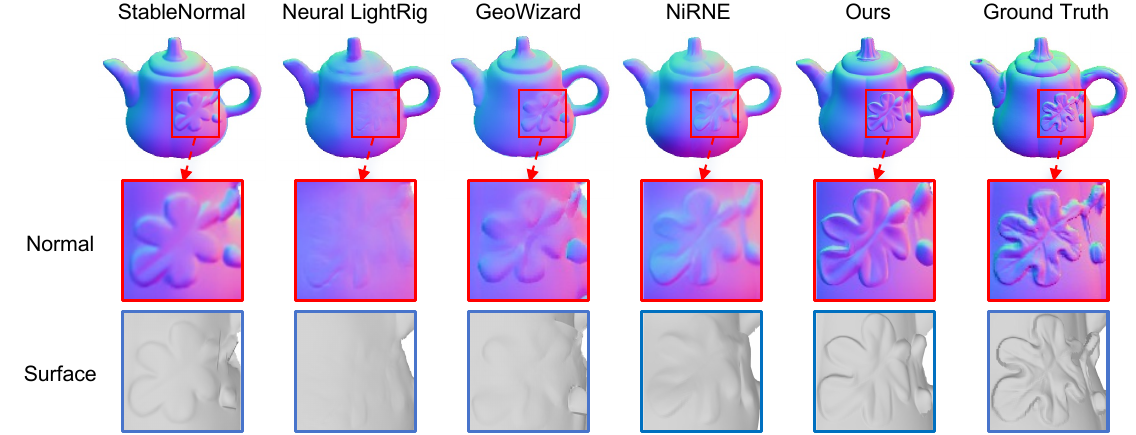}
    \caption{\textbf{Illustration of 3D misalignment.} The estimated normal maps of previous methods may appear to have an overall correct appearance, yet the reconstructed surfaces often fail to align with the accurate geometry details, showing over-smooth results. In contrast, our predicted normal maps exhibit better 3D alignment, leading to more faithful surface reconstruction. }
    \label{fig:3d_misalignment}
    \vspace{-10pt}
\end{figure}

\section{Introduction}
Normal maps represent 3D geometry by specifying the surface orientation at each pixel point, making them essential for a wide range of applications, including relighting~\citep{tiwari2024merlin, sang2020single, li2018learning, liu2020unsupervised, li2023dani}, 3D  reconstruction~\citep{ye2025hi3dgen}, and rendering~\citep{zeng2025survey,rendering2015physically}. 
In early methods, obtaining accurate normal maps required specialized hardware setups and was therefore costly for broad use. This motivates the development of techniques for monocular normal estimation, which aim to reliably infer normal maps directly from casually captured RGB images.

\begin{wrapfigure}{r}{0.38\textwidth}
    \vspace{-10pt}
  \centering
  \includegraphics[width=0.37\textwidth]{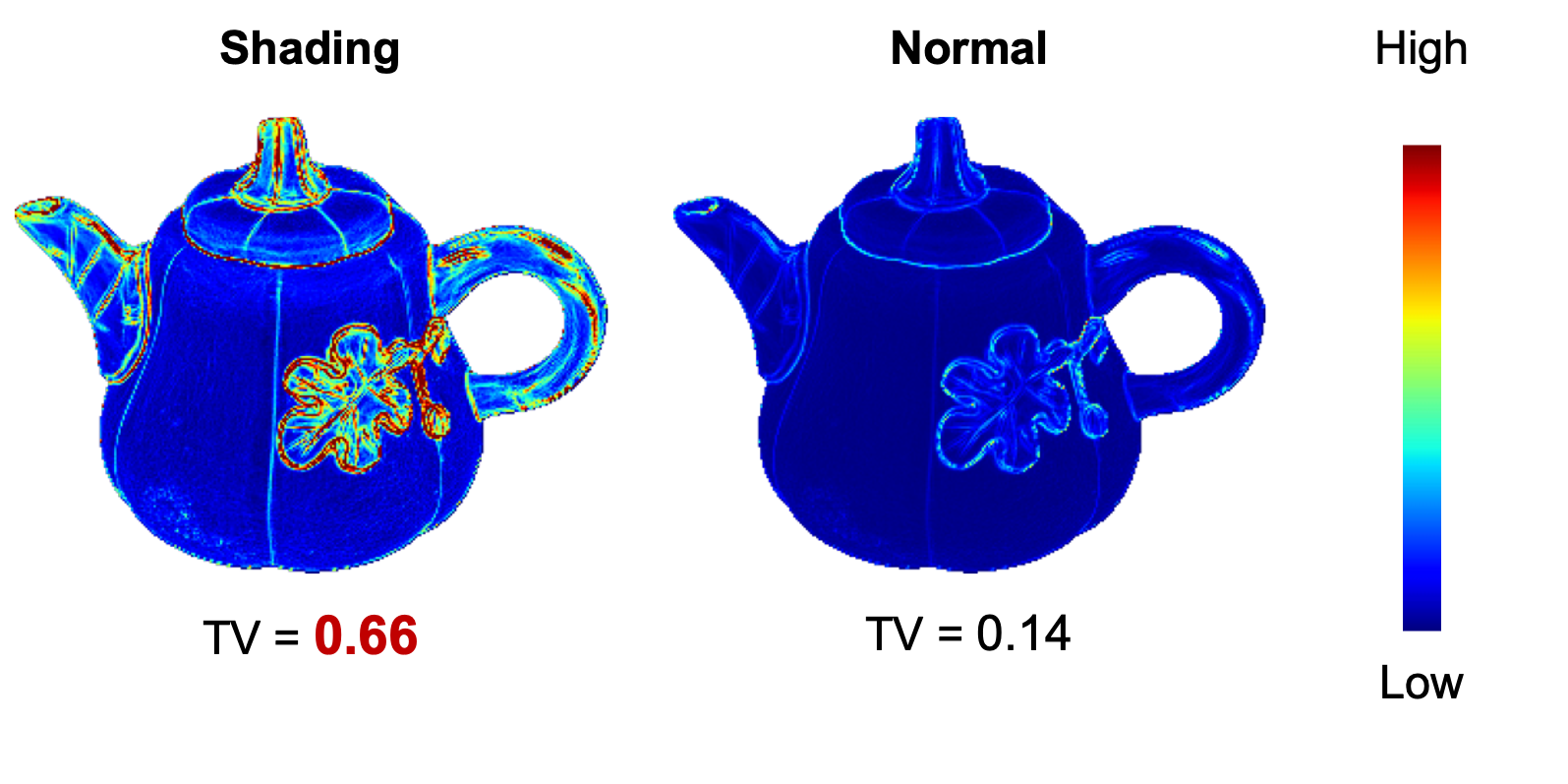}
  \caption{\textbf{Validation of sensitivity to geometry variations for different representations}, including the proposed shading sequence (left) and the normal map (right), measured by average total variation (TV). TV is computed as the mean magnitude of the first-order image's gradient in terms of different representations, where higher TV indicates stronger sensitivity to spatial geometric variation. }
  \label{fig:TV}
  \vspace{-13pt}
\end{wrapfigure}

Specifically, to achieve accurate monocular normal estimation, prior works~\citep{yoon2016fine, he2024neural, li2015depth, li2024spin, fu2024geowizard, bae2021estimating} directly predict normal maps from a single RGB image using deep neural models.
Despite achieving promising results, these methods often produce normal maps that appear to have a correct appearance but fail to remain consistent with the underlying 3D geometry. We refer to this limitation as ``\textbf{3D misalignment}'' (see \Fref{fig:3d_misalignment}).
We argue that this limitation arises from the current paradigm, where the model learns to recover geometry primarily by aligning with the color representation of normal maps. As a result, the deep model struggles to distinguish and estimate fine geometric details because normal maps represent geometry in a highly compact form, where geometry variations across different positions appear only as subtle color differences. This issue becomes particularly severe in monocular normal estimation, where geometric details in the input are inherently more ambiguous, thus difficult to recover.
 
To reduce 3D misalignment, this paper proposes a new paradigm for normal estimation built on a redesigned training target.
The main idea is to adopt a representation that is more sensitive to geometric variation as the ground-truth signal to be supervised, thereby enhancing the network’s ability to distinguish and estimate geometric details. Guided by this intuition, we propose using a \textbf{shading sequence}, defined as the clamped dot product between the normal map and a set of canonical light directions, as the new training target.
The idea of using a shading sequence is motivated by two key observations. First, shading sequences capture geometric variation through brightness variations, while excluding material influences, making them sensitive only to geometric variations, as illustrated in \Fref{fig:TV}. Second, predicting the shading sequence given canonical light directions is equivalent to predicting the normal map, as the shading sequence can be losslessly converted to the normal map~\citep{yu2010photometric} using an Ordinary Least Squares (OLS) solver. 

Based on the new paradigm, we propose \textbf{\ours}, a method that \textbf{R}eformulating normal estimation as the \textbf{S}hading sequence \textbf{E}stimation based on the monocular input image. Specifically, as the shading sequence can be regarded as a video, we leverage recent image-to-video generative models~\citep{voleti2024sv3d,blattmann2023stable} to predict shading sequences. A key benefit of this design is that the rich lighting priors encoded in large-scale video models facilitate the generation of accurate shading variations. Once the shading sequence is obtained, we recover the normal map using an OLS solver~\citep{woodham1980photometric}.
In practice, to enhance robustness when handling objects with more complex materials and lights, we train our model on a collected dataset named~\dataset, enriched with more diverse materials and light conditions compared to previous datasets. Experimental results demonstrate that our method achieves superior performance compared to state-of-the-art methods.
Overall, we summarize the main contributions of this paper as:  
\begin{itemize}
    \item We introduce a new paradigm that reformulates the task of monocular normal estimation as shading sequence estimation.
    \item Under the new paradigm, we propose \ours, a monocular normal estimation method using an image-to-video generative model that predicts a shading sequence of an object under predefined canonical parallel lights and analytically derives normal maps from it. 
    \item We train RoSE on~\dataset, a synthetic dataset with diverse material and light conditions. Experiments show that our method achieves state-of-the-art performance on multiple datasets, especially on the widely-used real-world benchmark datasets (\ie, DiLiGenT, LUCES).
\end{itemize}

\section{Related Works}
\noindent\textbf{Monocular Normal Estimation.} Despite persistent research efforts in monocular normal estimation, achieving high accuracy remains significantly challenging due to the complexity of this task, which requires the prediction of 3D geometry with highly limited input information. Early works~\citep{Eigen_2015_ICCV, do2020surface, fouhey2013data, wang2015designing, zhang2019pattern, bansal2016marr, ladicky2014discriminatively, li2015depth, wang2020vplnet} relied on handcrafted features, empirical priors, or conventional deep neural networks. However, these methods often suffer from limited generalization ability. Recent methods based on generative models~\citep{voleti2024sv3d, fu2024geowizard, he2024lotus}, physics-inspired deep networks~\citep{bae2024dsine}, and auto-regressive frameworks~\citep{ye2025hi3dgen} have demonstrated improved generalization ability and the capacity to estimate relatively accurate normal maps. 
However, the estimated normal maps suffer from 3D misalignment, a problem that stems from the current paradigm where the model fails to capture the compact information in the normal maps.
To address this, other works~\citep{tiwari2024merlin, he2024neural} attempt to first generate more input images under a set of predefined canonical lights and subsequently estimate normals from these multi-light images. Yet, the accuracy of such methods is often degraded by artifacts in the generated input images, resulting in more severe 3D misalignment. In contrast, we propose a new paradigm that uses shading sequences, a representation that is sensitive to geometry, as the training target, and leverage video generative models to estimate them based on the input image, which achieves improved 3D alignment.

\textbf{Video Generation.} Recent advances in video generation~\citep{rombach2022high, peebles2023scalable, zhang2023adding, ho2022video} have significantly improved video synthesis quality. Methods such as~\citep{blattmann2023stable,deng2024autoregressive,kuaishou,lin2024open,guo2023animatediff} can generate high-fidelity videos by enforcing temporal consistency across frames through deep architectures such as temporal UNets~\citep{blattmann2023stable,guo2023animatediff} and Transformers~\citep{deng2024autoregressive,lin2024open}.
In 3D generation, video diffusion models are used to facilitate cross-view consistency~\citep{voleti2024sv3d, tang2024lgm, dai2023animateanything} to improve the quality of generated 3D models.
\blue{In 3D estimation, recent work~\citep{bin2025normalcrafter} employs a video diffusion model for normal estimation. 
They focus on predicting per-frame normals for an input video. In contrast, our work leverages the capability of video generative models to predict a shading sequence that follows a predefined light path consisting of multiple canonical parallel lights, using only a single input image. This enables accurate monocular normal estimation for objects with diverse shapes and materials.}

\textbf{Shading Utilization.} In photometric stereo methods, shadings are often used to explain the behavior of a deep model in normal map estimation. Previous studies~\citep{chen2020learned,wei2025revisiting} show that learned features closely resemble shading sequences, motivating the use of shading sequence supervision to improve network performance~\citep{wei2025revisiting}. Inspired by these findings, we reformulate monocular normal estimation as a shading sequence estimation task and train a video diffusion model with shading sequences as targets.

\section{Methods}
\subsection{On Equivalence of Normal Estimation and Shading Sequence Estimation}
\label{sec:rendering_equation}

\noindent \textbf{Shading map and shading sequence}. In this paper, we define shading map~\citep{wei2025revisiting} as:
\begin{equation}
\mathbf{S} \triangleq \{\mathbf{s}_\mathrm{p}=\max(\mathbf{n}_\mathrm{p} \cdot \mathbf{l}, 0)|\mathrm{p}\in \mathcal{P}\},    
\end{equation}
where $\mathbf{n}$ is the normal map, and $\mathbf{l}$ is the direction of parallel light, \blue{$\max(.,0)$ is the nonlinear maximum operation that clamp the negative values}, $\mathcal{P}$ is the points that belong to the object. Shading maps remove the effects of surface reflectance and occlusion-induced cast shadows while preserving the geometry information and \blue{the attached shadow}. A sequence of shading maps obtained under multiple canonical lights, defined as a \textbf{shading sequence}, $\mathbf{S}^{s} \triangleq \{\mathbf{S}_i \mid i \in {1, \dots, f}\}$, provides sensitive cues to the underlying 3D geometry.

\textbf{Normal map estimation}. Given an observed image $\mathbf{I}$ of an object captured under arbitrary lights, the goal of monocular normal estimation is to recover the normal map $\mathbf{N}\triangleq\{\mathbf{n}_\mathrm{p}|\mathrm{p}\in \mathcal{P}\}$. This requires learning a mapping:
\begin{equation}
\Phi: \mathbf{I} \rightarrow \mathbf{N}.
\end{equation}
Previous methods rely on deep models to learn a direct color mapping from a single RGB image to a normal map. This often produces a visually aligned appearance but inaccurate 3D geometry (\ie, the 3D misalignment). A more recent line of work~\citep {he2024neural,tiwari2024merlin} first generates a series of RGB images under simple light sources and then estimates normals from them. The main idea of these works is to augment the input with additional generated images that provide additional cues, thereby improving the prediction of normal maps. However, as the materials, lights, and geometry in the input image become more complex, the process of generating additional RGB images itself introduces substantial bias, ultimately leading to more pronounced 3D misalignment artifacts.

\textbf{Shading sequence estimation}.
The shading sequence under a set of predefined, non-coplanar, parallel lights (canonical lights) can be converted into a normal map. This enables us to shift the training target to shading sequence prediction with lights that vary along a predefined path, $\mathbf{L} \triangleq \{\mathbf{l}_i \mid i \in {1, \dots, f}\}$.
\begin{equation}
\Phi_S: \mathbf{I}_g \rightarrow \mathbf{S}^{s},
\end{equation}
where $\mathbf{I}_g$ denotes the grayscale input image. Then, the shading-to-normal estimation $\mathbf{S}^s \rightarrow \mathbf{N}$ can be solved via Ordinary Least Squares (OLS)~\citep{woodham1980photometric}:
\begin{align}
\label{eq:ls}
\mathbf{N} = \arg\min_\mathbf{N}\|\mathbf{N}^\top \mathbf{L}-\mathbf{S}^s\|^2
= (\mathbf{L}^\top\mathbf{L})^{-1}\mathbf{L}^\top \mathbf{S}^s.
\end{align}
\blue{The solution is unique when $\mathbf{L}$ is full rank. In practice, however, the $\max(\cdot, 0)$ operation causes a truncation effect, which makes the OLS estimate biased if OLS is applied directly to the shading sequence. To address this issue, we solve for the normal using only shadings with values greater than 0, treating them as valid equations in OLS.}
\subsection{Shading Sequence-based Training Target}
\begin{wrapfigure}{r}{0.38\textwidth}
    \vspace{-10pt}
  \centering
  \includegraphics[width=0.37\textwidth]{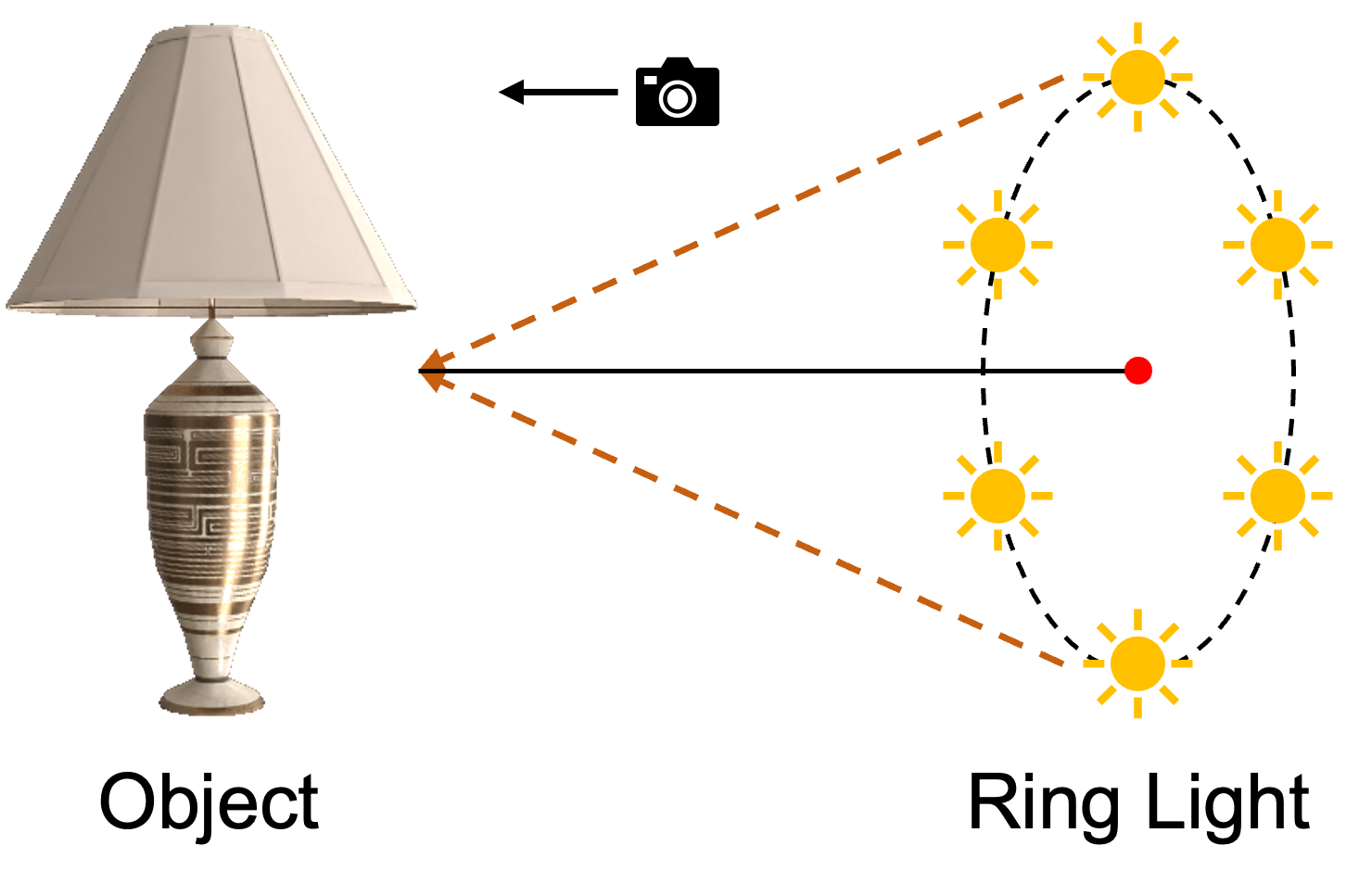}
  \caption{\textbf{Ring light setup}. }
  \label{fig:ring-light}
  \vspace{-20pt}
\end{wrapfigure}

Reformulating monocular normal estimation as shading sequence estimation introduces additional flexibility in designing the training target, since different choices of $\mathbf{L}$ yield different shading sequences. As long as each surface point is illuminated by at least three non-coplanar parallel light sources (\ie, the lighting matrix $\mathbf{L}$ is full rank in \Eref{eq:ls}), normal maps can be recovered from the shading sequence without information loss. 
In our setup, this means that each surface point should correspond to at least three positive shading values. 
In this paper, we adopt a classic ring light setup from photometric stereo~\citep{zhou2010ring}, where canonical lights are uniformly placed on a latitude ring in the upper hemisphere of the object's surface, each light oriented toward the surface center (see \Fref{fig:ring-light}). 
With an appropriate choice of the ring’s latitude (\blue{$45^\circ$ in our setup}), these lights collectively illuminate all surface points. The remaining question is: {\it How many lights with distinct positions do we need, in particular the minimum number $l_{\min}$, to guarantee that every surface point is illuminated by at least three light sources with positive shading values?} We address this in Lemma~\ref{lemma:light}.

\begin{lemma}
\label{lemma:light}
Define a surface point as illuminated if its shading value is positive, \ie, $\max(0, \mathbf{S}) > 0$. Under a single parallel light, at least half of the upper hemisphere is illuminated. Therefore, $n=2$ lights are sufficient to ensure that every point on the hemisphere is illuminated once. To further guarantee that every point is illuminated by at least $m=3$ different lights, the minimum number of lights is $l_{\min} = m \times n = 6$, with the light directions uniformly distributed along the ring.
\end{lemma}

In our experiments, we observed that slightly increasing the number of light sources improved the accuracy of both normal and shading estimation. The best performance was achieved with 9 light sources, yielding a $0.74^\circ$ improvement on LUCES~\cite{mecca2021luces} compared to 6 lights. While appropriately introducing additional light sources can further enhance accuracy, it also incurs longer training and convergence time as well as higher resource consumption. For instance, under the same settings, performance dropped by $1.31^\circ$ under 12 lights.

\begin{figure}[t]
    \centering
    \includegraphics[width=\textwidth]{"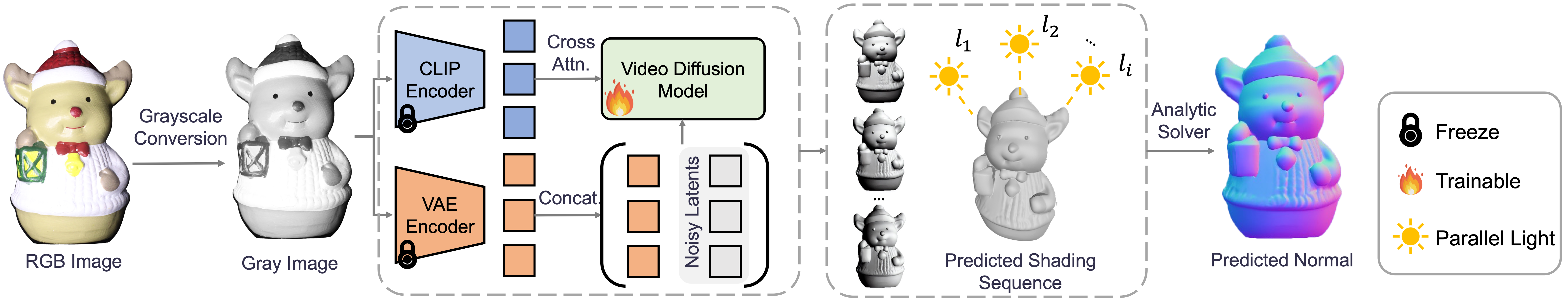"}
    \caption{\textbf{Pipeline of~\ours.} Given a monocular RGB image under arbitrary light, \ours~first converts it into a grayscale image, which is then used to generate the shading sequence via a video diffusion model. This generation is guided by two complementary feature representations extracted from a CLIP encoder and a VAE encoder. Finally, an ordinary least squares problem is solved analytically to estimate the normal map from the generated shading sequence. We train the video diffusion model while freezing the CLIP and the VAE encoder.} 
    \label{fig:pipeline}
    \vspace{-0.1in}
\end{figure}
\subsection{\ours: a Monocular Normal Estimator based on Video Generative Model}
\label{sec:mlvsd}

\textbf{Architecture of~\ours}~is shown in \Fref{fig:pipeline}. Firstly, the shading generator $\mathbf{g}_\theta(\cdot)$ is designed to take grayscale images $\mathbf{I}_g$ as input, effectively eliminating redundant chromatic information that may distract the model from learning geometric cues. It produces grayscale shading sequences that follow a predefined light path, introducing structured patterns and temporal coherence well-suited to video generation models.
In this paper, we implement $\mathbf{g}_\theta(\cdot)$ using a standard video diffusion U-Net composed of multiple spatial and temporal transformer blocks~\citep{voleti2024sv3d, blattmann2023stable}. The grayscale input image $\mathbf{I}_g$ is used as an additional condition to guide the denoising process during shading sequence generation.

Specifically, following previous works~\citep{he2024neural, voleti2024sv3d, blattmann2023stable}, we adopt a similar dual-branch conditioning strategy that combines global guidance from CLIP embedding and local guidance from VAE latent concatenation to reuse the pre-trained weights of the model.
1) \textbf{CLIP embedding}: We extract a global feature vector $\mathbf{c}_g$ from the input image using a pretrained CLIP encoder. This semantic embedding is injected into the denoising U-Net via cross-attention, guiding the shading generation with object-level context.
2) \textbf{VAE latent concatenation}: to preserve spatial details, we encode the grayscale input $\mathbf{I}_G$ with a pretrained VAE encoder $\mathcal{E}$ and concatenate the resulting latent with the noisy latent $z_t$ at each denoising step.
Since $\mathbf{I}_G$ is single-channel, we replicate it to three channels before feeding it into the VAE and CLIP encoders: $\mathbf{I}_g^\prime=\operatorname{repeat}(\mathbf{I}_g, B\times H \times W \rightarrow B\times H \times W \times 3)$,
where $B$ is the batch size. By combining both conditioning techniques, the model better preserves the fine geometric structures of the input image during generation, which is particularly important for accurate normal estimation.
Finally, the generated latents are decoded by the VAE decoder $\mathcal{D}$ and averaged across channels to obtain the final grayscale shading sequence.

\textbf{Training.} During training, we use the standard training objectives on latent space encoded by $\mathcal{E}$. The video generative model will learn to predict the noise given the noisy latent $\mathbf{z}_t$, $\mathbf{z}_0=\mathcal{E}(\mathbf{S^s})$, where $\mathbf{z}_t =\alpha_t \epsilon+\sigma_t \mathbf{z}_0$. The diffusion loss follows calculation of $\mathbf{z}_0$-reparameterization~\citep{ho2020denoising}: 
\begin{equation}
\mathcal{L}_{\mathrm{diff}}=\mathbb{E}_{\mathbf{z}_0, c, t}\left\|\mathbf{z}_0-\hat{\mathbf{z}}_0\right\|^2, \hat{\mathbf{z}}_0 =\frac{\mathbf{z}_t-\alpha_t\mathbf{g}_\theta\left(\mathbf{z}_t |c^\prime, t\right)}{\sigma_t}.
\end{equation}
where $\hat{\mathbf{z}}_0$ the one-step denoised version of ${\mathbf{z}}_t$. 

\textbf{Dataset Curation}. To improve robustness when handling more complex materials and lights, we curate a dataset named~\dataset, featuring diverse shapes, materials, and light conditions to ensure robust generalization. 
\dataset~is built upon a list of pre-filtered 3D models (90K) curated from Objaverse~\citep{deitke2023objaverse,he2024neural}, a widely adopted resource for 3D generation and reconstruction.
For each object, we render observed images under three lighting setups: (1) \textbf{parallel lights} randomly placed around the object; (2) \textbf{point lights} with randomly sampled positions and intensities; and (3) \textbf{environment lights} using high-dynamic-range (HDR) maps sampled from a public collection of 780 real-world environments~\citep{polyheaven}. Each object is rendered from six distinct viewpoints (top, left, right, bottom, front, and one random view) to ensure comprehensive geometric coverage. To avoid lighting from the object’s backside in each view, we apply view-dependent transformations to keep light sources in the upper hemisphere relative to the view direction.  
During rendering, we implement \textbf{material augmentation} to the dataset by either retaining the object’s original texture or applying material augmentation. With a probability of $0.5$, an additional material is assigned from the MatSynth dataset~\citep{vecchio2024matsynth}, which contains 5,657 high-quality PBR materials. More specifically, we assign a probability of $0.25$ to sample materials from the metallic category, while $0.25$ to extract materials from non-metallic categories such as plastic, wood, and fabric. This augmentation improves surface diversity and model robustness, especially for metallic materials.
All images are rendered using Blender at a resolution of $576 \times 576$ following~\citep{voleti2024sv3d}, generating approximately 3 million image-normal pairs. Precomputed shading sequences under known canonical light sources are also provided. More details on rendering parameters, camera setup, and augmentation strategies are in the appendix.

\section{Experiments}
\label{sec-exp}
\subsection{Experiment setup}
\textbf{Datasets}.
We evaluate the proposed method on widely used benchmarks, including LUCES~\citep{mecca2021luces} for near-light monocular normal estimation, DiLiGenT~\citep{shi2016benchmark} for parallel-light settings, and a curated test set of 100 unseen objects from the Objaverse dataset~\citep{deitke2023objaverse} rendered with diverse materials and light conditions. 

\textbf{Baselines.} We compare \ours~with 7 other monocular normal estimation methods, \ie, GeoWizard~\citep{fu2024geowizard}, DSINE~\citep{bae2024dsine}, StableNormal~\citep{ye2024stablenormal}, Lotus-G \& Lotus-D~\citep{he2024lotus}, Neural LightRig~\citep{he2024neural}, and NiRNE~\citep{ye2025hi3dgen}.

\noindent\textbf{Implementation details and evaluation metrics.} All training experiments are conducted on 8 $\times$ NVIDIA H100 GPUs with 80GB memory. The model is trained at a learning rate of $1\times e^{-5}$, using AdamW as the optimizer. The diffusion architecture follows previous work~\citep{voleti2024sv3d}. More details can be found in the appendix. To assess the accuracy of predicted normal maps, following the common protocol in previous works~\citep{ye2024stablenormal,ye2025hi3dgen,bae2024dsine,he2024lotus}, we use the mean angular error (MAE) as the evaluation metrics for all experiments.

\begin{table*}[t]
    \setlength{\tabcolsep}{6pt}
    \caption{Quantitative comparison in terms of MAE ($\downarrow$) of the normal map on DiLiGenT benchmark dataset. Highlighted numbers indicate the \colorbox{LightGreen!40}{best} and \colorbox{LightSkyBlue!40}{second best} results among monocular estimation methods.}
    \centering
    \label{tab:diligent_quan}
    \resizebox{\textwidth}{!}{
    \footnotesize
    \begin{NiceTabular}{l|cccccccccc|c}
    \toprule
    Method & {\sc Ball} & {\sc Bear} & {\sc Buddha} & {\sc Cat} & {\sc Cow} & {\sc Goblet} & {\sc Harvest} & {\sc Pot1} & {\sc Pot2} & {\sc Reading} & Mean \\
    \midrule
    GeoWizard & 16.85 & 14.58 & 26.38 & 21.82 & 19.54 & 17.70 & 29.78 & 21.86 & 19.97 & 29.42 & 21.79 \\
    DSINE & 23.82 & 14.15 & 28.09 & 18.22 & 19.35 & 22.63 & 35.90 & 20.90 & 19.14 & 30.25 & 23.25 \\
    StableNormal & 17.11 & 13.17 & 21.84 & 22.46 & 22.63 & 15.96 & 32.14 & 17.43 & 16.53 & 25.15 & 20.44 \\
    Lotus-D & 36.83 & 11.29 & 21.68 & 23.93 & 22.62 & \cellcolor{LightGreen!40} 13.93 & 34.99 & 21.45 & 17.14 & 25.49 & 22.94 \\
    Lotus-G & 12.74 & 13.02 & 23.27 & 22.68 & 22.78 & \cellcolor{LightSkyBlue!40} 15.52 & 32.94 & 23.27 & 19.23 & 28.67 & 21.41 \\
    Neural LightRig & \cellcolor{LightSkyBlue!40} 10.16 & 14.47 & 26.23 & 28.39 & 21.16 & 22.70 & 76.82 & 24.71 & 31.84 & 34.51 & 29.10 \\
    NiRNE & 10.26 & \cellcolor{LightSkyBlue!40} 10.87 & \cellcolor{LightSkyBlue!40} 21.28 & \cellcolor{LightGreen!40} 15.43 & \cellcolor{LightSkyBlue!40} 15.03 & 17.91 & \cellcolor{LightGreen!40} 27.40 & \cellcolor{LightGreen!40} 15.27 & \cellcolor{LightSkyBlue!40} 16.15 & \cellcolor{LightGreen!40} 23.08 & \cellcolor{LightSkyBlue!40} 17.27 \\
    \midrule
    \textbf{Ours} & \cellcolor{LightGreen!40} \textbf{5.51} & \cellcolor{LightGreen!40} \textbf{9.22} & \cellcolor{LightGreen!40} \textbf{20.72} & \cellcolor{LightSkyBlue!40} \textbf{15.78} & \cellcolor{LightGreen!40} \textbf{13.28} & \textbf{16.55} & \cellcolor{LightSkyBlue!40} \textbf{28.62} & \cellcolor{LightSkyBlue!40} \textbf{16.05} & \cellcolor{LightGreen!40} \textbf{14.24} & \cellcolor{LightSkyBlue!40} \textbf{23.65} & \cellcolor{LightGreen!40} \textbf{16.36} \\
    \bottomrule
    \end{NiceTabular}
        }
\vspace{-5pt}
\end{table*}

\begin{table*}[t]
    \setlength{\tabcolsep}{4pt}
    \caption{Quantitative comparison in terms of MAE of the normal map on LUCES benchmark dataset~\citep{mecca2021luces}. Highlighted numbers indicate the \colorbox{LightGreen!40}{best} and \colorbox{LightSkyBlue!40}{second best} results among monocular estimation methods.}
    \centering
    \label{tab:luces_quan}
    \resizebox{\textwidth}{!}{
    \footnotesize
    \begin{NiceTabular}{l|cccccccccccccc|c}
        \toprule
        Method & {\sc Ball} & {\sc Bell} & {\sc Bowl} & {\sc Buddha} & {\sc Bunny} & {\sc Cup} & {\sc Die} & {\sc Hippo} & {\sc House} & {\sc Jar} & {\sc Owl} & {\sc Queen} & {\sc Squirrel} & {\sc Tool} & Mean \\
        \midrule
        GeoWizard  & 30.09 & 9.08 & 22.29 & 22.71 & 15.90 & 20.20 & 15.76 & 17.55 & 42.15 & 11.07 & 28.68 & 25.36 & 35.48 & 18.57 & 22.49 \\
        DSINE & 26.88 & 15.00 & \cellcolor{LightSkyBlue!40} 9.53 & 22.34 & 15.82 & 22.65 & 32.02 & \cellcolor{LightSkyBlue!40} 14.42 & 36.95 & 16.26 & 27.46 & 23.76 & 25.26 & 17.19 & 21.82 \\
        StableNormal & 9.58 & 9.36 & 31.39 & 20.80 & 14.73 & 29.40 & \cellcolor{LightSkyBlue!40} 11.88 & 20.80 & 37.55 & \cellcolor{LightSkyBlue!40} 8.25 & 23.23 & 21.10 & 27.24 & 19.49 & 20.34 \\
        Lotus-D & 17.94 & 9.50 & 11.43 & \cellcolor{LightSkyBlue!40} 19.70 & 12.99 & 37.44 & 13.14 & 15.85 & \cellcolor{LightSkyBlue!40} 35.30 & 9.69 & 20.53 & \cellcolor{LightGreen!40} 19.72 & 23.52 & 13.15 & 18.56 \\
        Lotus-G & 17.82 & \cellcolor{LightSkyBlue!40} 8.66 & 10.89 & 19.71 & \cellcolor{LightSkyBlue!40} 12.90 & 23.26 & 12.59 & 16.94 & 35.32 & 10.69 & \cellcolor{LightGreen!40} 18.94 & 20.65 & 24.05 & \cellcolor{LightSkyBlue!40} 11.74 & \cellcolor{LightSkyBlue!40} 17.44 \\
        Neural LightRig & \cellcolor{LightSkyBlue!40}9.52 & 11.95 & 21.71 & 20.66 & 15.25 & 18.08 & 25.13 & 18.54 & 39.67 & 19.78 & 23.40 & 23.35 & 25.32 & 20.97 & 20.95 \\
        NiRNE  & 10.55 & 12.00 & 17.35 & 20.62 & 16.14 & \cellcolor{LightSkyBlue!40} 15.78 & 12.57 & 15.85 & \cellcolor{LightGreen!40} 34.99 & 10.37 & 22.46 & 22.41 & \cellcolor{LightSkyBlue!40} 21.90 & 17.34 & 17.88 \\
        \midrule
        \textbf{Ours} & \cellcolor{LightGreen!40} \textbf{9.09} & \cellcolor{LightGreen!40} \textbf{5.94} & \cellcolor{LightGreen!40} \textbf{6.84} & \cellcolor{LightGreen!40} \textbf{17.58} & \cellcolor{LightGreen!40} \textbf{12.70} & \cellcolor{LightGreen!40} \textbf{13.80} & \cellcolor{LightGreen!40} \textbf{8.26} & \cellcolor{LightGreen!40} \textbf{14.14} & \textbf{36.79} & \cellcolor{LightGreen!40} \textbf{5.93} & \cellcolor{LightSkyBlue!40} \textbf{19.60} & \cellcolor{LightSkyBlue!40} \textbf{19.99} & \cellcolor{LightGreen!40} \textbf{21.34} & \cellcolor{LightGreen!40} \textbf{10.66} & \cellcolor{LightGreen!40} \textbf{14.48}\\
        \bottomrule
    \end{NiceTabular}    }
    \vspace{-10pt}
\end{table*}

\subsection{Performance on Benchmark Dataset}
We conduct monocular normal estimation experiments on the DiLiGenT~\citep{shi2016benchmark} and LUCES benchmark dataset~\citep{mecca2021luces} to evaluate our method’s ability in handling objects captured under distant and near-field light sources. For each object, we select 10 images with relatively centered lights so that the light can cover enough details. The index of the images used for testing can be found in the appendix.

\begin{figure}[t]
    \centering
    \includegraphics[width=\textwidth]{"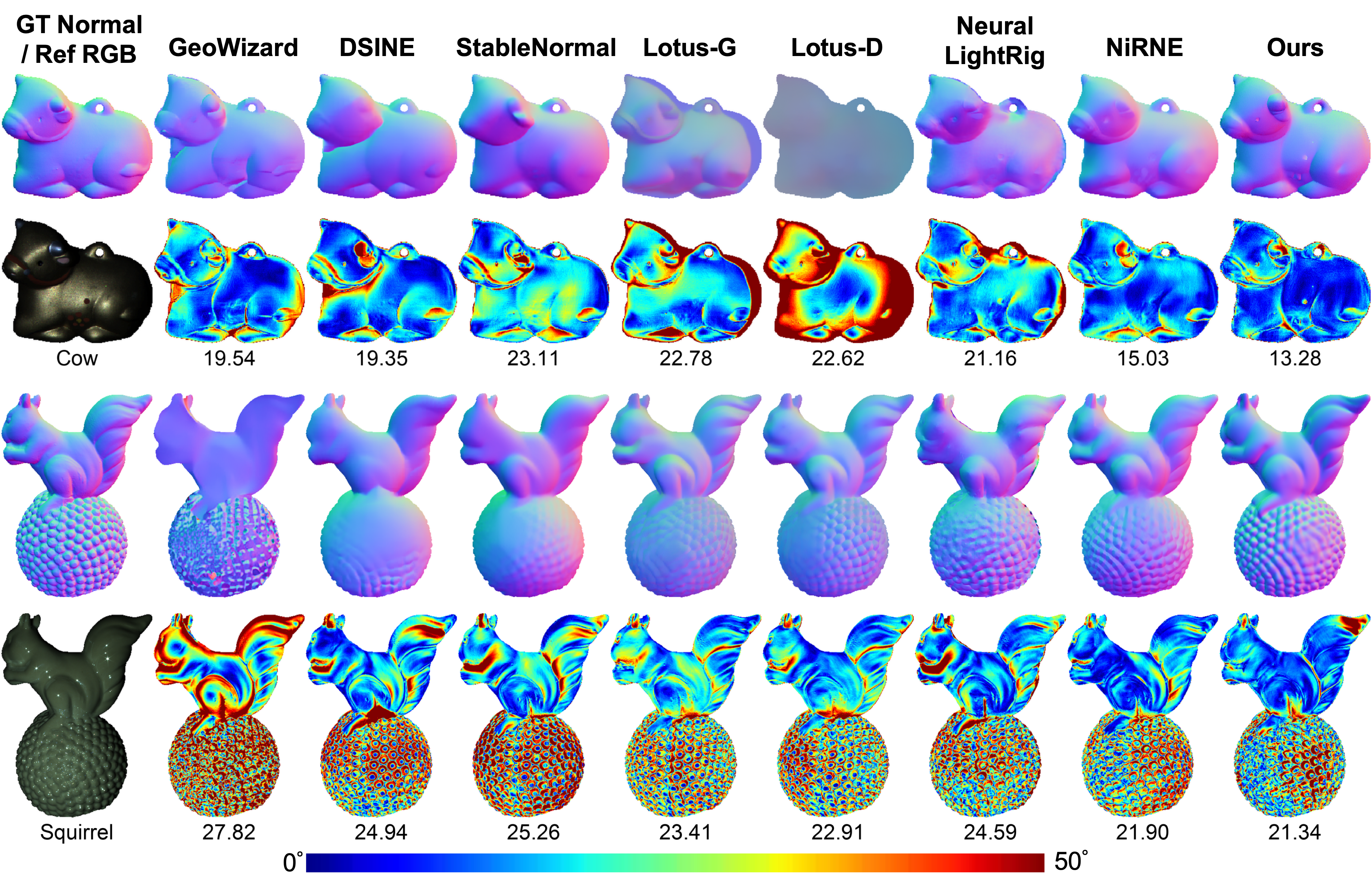"}
    \caption{\textbf{Qualitative comparison} on selected objects from two benchmark dataset ({\sc Cow} from {\sc DiLiGenT}~\citep{shi2016benchmark} and {\sc Squirrel} from {\sc LUCES}~\citep{mecca2021luces}. Row 1 \& 3: normal map comparison. Row 2 \& 4: error map comparison.)
    Best viewed in color with zooming in.} 
    \label{fig:benchmark_quali}
    \vspace{-10pt}
\end{figure}

\begin{table*}[t]
\setlength{\tabcolsep}{4mm}
    \centering
    \caption{Quantitative comparison in terms of \textbf{Mean} and \textbf{Median} Angular Errors of the normal map on \dataset~test set, and the percentage of objects below a specific error bound. Highlighted numbers indicate the \colorbox{LightGreen!40}{best} and \colorbox{LightSkyBlue!40}{second best} results among monocular estimation methods.}
    \label{tab:method_comparison}
    \footnotesize
    \resizebox{\textwidth}{!}{
    \begin{NiceTabular}{l|cc|cccccc}
        \toprule
        Method & Mean $\downarrow$ & Median $\downarrow$ & $3^{\circ}$ $\uparrow$ & $5^{\circ}$  $\uparrow$& $7.5^{\circ}$ $\uparrow$ & $11.25^{\circ}$ $\uparrow$ & $22.5^{\circ}$ $\uparrow$ & $30^{\circ}$ $\uparrow$\\
        \midrule
        GeoWizard &20.46 &11.61 &12.84 &25.41 &37.34 &49.09 &68.53 &76.29\\
        DSINE &22.53 &14.04 &12.38 &22.47 &32.18 &43.27 &65.19 &74.16\\
        StableNormal &19.71 &11.23 &6.83 &18.67 &34.65 &50.08 &71.66 &79.48\\
        Lotus-D & \cellcolor{LightSkyBlue!40} 18.48 & \cellcolor{LightSkyBlue!40} 10.63 &14.51 &26.34 &38.78 & \cellcolor{LightSkyBlue!40} 51.78 & \cellcolor{LightSkyBlue!40} 72.47 &79.82\\
        Lotus-G &18.76 &10.65 &14.67 &27.13 & \cellcolor{LightSkyBlue!40} 39.19 &51.63 &71.83 &79.54\\
        Neural LightRig &20.59 &11.36 & \cellcolor{LightSkyBlue!40} 17.65 & \cellcolor{LightSkyBlue!40} 27.59 &37.90 &49.69 &70.85 &78.54 \\
        NiRNE &19.57 &13.57 &4.06 &11.92 &25.53 &42.10 &71.42 & \cellcolor{LightSkyBlue!40} 81.21\\
        \midrule
        \textbf{Ours} & \cellcolor{LightGreen!40} \textbf{15.37} & \cellcolor{LightGreen!40} \textbf{7.78} & \cellcolor{LightGreen!40} \textbf{26.99} & \cellcolor{LightGreen!40} \textbf{38.38} & \cellcolor{LightGreen!40} \textbf{49.00} & \cellcolor{LightGreen!40} \textbf{60.32} & \cellcolor{LightGreen!40} \textbf{78.30} & \cellcolor{LightGreen!40} \textbf{84.28}\\
        \bottomrule
    \end{NiceTabular}
}
\vspace{-15pt}
\end{table*}

\noindent\textbf{Quantitative analysis \blue{on normal estimation}}.
The results in \Tref{tab:diligent_quan} and \Tref{tab:luces_quan} present the average MAE for each object across selected 10 images, and the average MAE across all objects.
These quantitative results demonstrate a significant advantage of our method over the state-of-the-art method ($16.36^\circ$ for ours vs. $17.27^\circ$ for NiRNE~\citep{ye2025hi3dgen} on DiLiGenT dataset~\citep{shi2016benchmark}; $14.48^\circ$ for ours vs. $17.44^\circ$ for Lotus-G~\cite{he2024lotus} on LUCES dataset~\citep{mecca2021luces}). This validates the effectiveness of our method in achieving robust performance in normal estimation across various materials and light conditions. However, we observe that for certain objects, such as {\sc Goblet} in DiLiGenT and {\sc House} in LUCES, our method does not rank within the top two. We attribute this to the model's inherent variance. Note that even the previous SOTA method, NiRNE, fails to deliver consistently strong performance across all cases. Another possible reason may be attributed to the training set used. We discuss this in~\Sref{sec:abl}.

\noindent\textbf{Qualitative analysis \blue{on normal estimation}}.
We present a qualitative comparison between our method and state-of-the-art methods in~\Fref{fig:benchmark_quali}.
Our method consistently recovers accurate object details in the estimated normal maps, achieving lower MAE. In contrast, previous methods often produce overly smooth results, distorted normal distributions, or significant artifacts~\citep{he2024neural} (\eg, the tail and back of the {\sc Squirrel}). These results validate the effectiveness of the proposed shading sequence-based formulation and demonstrate \ours’s advantage in preserving fine shape details for accurate normal estimation.

\blue{\noindent\textbf{Analysis on shading sequence estimation}. 
We conduct quantitative analyses\footnote{\blue{Please refer to the appendix for qualitative analyses.}} of the predicted shading sequences on the LUCES dataset to illustrate RoSE’s ability to recover accurate shading sequences. The shading map of all other methods (including the ground truth) is computed as the dot product between the lights' directions and the surface normal, with negative values clamped to zero. We use PSNR ($\uparrow$), SSIM ($\uparrow$), and LPIPS ($\downarrow$) as the evaluation metrics, as shown in~\Tref{tab:quan_shad}. The results demonstrate that \ours~achieves SOTA performance in predicted shading sequence, which also align with the results of normal estimation. }
\begin{table}[t]
\centering
\caption{\blue{Quantitative comparison on estimated shading sequence in terms of PSNR ($\uparrow$), SSIM ($\uparrow$), and LPIPS ($\downarrow$) on LUCES benchmark dataset~\citep{mecca2021luces}. 
Highlighted numbers indicate the \colorbox{LightGreen!40}{best} and \colorbox{LightSkyBlue!40}{second best} results.}}
\label{tab:quan_shad}
\resizebox{\textwidth}{!}{
\begin{tabular}{l|cccccccc}
\toprule
Metrics & GeoWizard & DSINE & StableNormal & Lotus-D & Lotus-G & Neural LightRig & NiRNE & \textbf{Ours} \\
\midrule
PSNR ($\uparrow$)
& 16.86 & 17.05 & 18.40 & 18.80 & \cellcolor{LightSkyBlue!40}19.19 & 17.88 & 18.99 
& \cellcolor{LightGreen!40}\textbf{20.74} \\
SSIM ($\uparrow$) 
& 0.6920 & 0.7199 & 0.7411 & 0.7492 & \cellcolor{LightSkyBlue!40}0.7589 & 0.7139 & 0.7503 
& \cellcolor{LightGreen!40}\textbf{0.7744} \\
LPIPS ($\downarrow$)
& 0.2806 & 0.3100 & 0.2972 & 0.2868 
& 0.2724 
& 0.2831 & \cellcolor{LightSkyBlue!40}0.2688
& \cellcolor{LightGreen!40}\textbf{0.2583} \\
\bottomrule
\end{tabular}
}
\end{table}

\subsection{Performance on~\dataset}
To further evaluate our method's performance across various lighting conditions and materials, the test set of the applied synthetic dataset consists of 100 unseen objects from Objaverse~\citep{deitke2023objaverse}. Each object is rendered with random materials selected from the MatSynth test set~\citep{vecchio2024matsynth}. For lighting conditions, we employ one random point light, one directional (parallel) light, and two environmental lights selected from Poly Haven~\citep{polyheaven} that are different from the training dataset. Each object is rendered from seven viewpoints, including the front, back, left, right, and top views, as well as two randomly sampled views. This setup yields a total of 2800 test samples. Following the evaluation protocol in prior work~\citep{he2024neural}, we report the mean and median angular error (MAE) across all objects, as well as the percentage of objects with MAE below specified angular thresholds. As shown in~\Tref{tab:method_comparison}, our method consistently outperforms baseline approaches across all metrics, with particularly strong performance under tighter thresholds (\ie, $3^\circ$-$7.5^\circ$), highlighting the robustness and accuracy of the proposed \ours.

\subsection{Ablation Study}
\label{sec:abl}
We conduct ablation experiments using the LUCES benchmark dataset~\citep{mecca2021luces} as the test set to analyze the effectiveness of the proposed~\ours~and~\dataset. 
Additional experiments, analysis, and discussion are in the appendix.

\textbf{Validation on details alignment.} Following~\citep{ye2025hi3dgen}, we evaluate detailed alignment by computing the sharp normal error (SNE), \ie, the normal estimation error measured in boundary regions, on the estimated normal maps from the LUCES dataset, see~\Tref{tab:abl_detail}. Our method achieves performance comparable to the state of the art method (NiRNE~\cite{ye2025hi3dgen}) and shows a clear advantage over other baselines. It is worth noting that NiRNE was trained on a dataset nearly $10\times$ larger and containing significantly more diverse and complex 3D models than ours. These results highlight that the proposed \ours~is capable of generating fine-grained details even with substantially lower resource consumption during training.

\textbf{Validation on negative-clamping on shading sequence.} After clamping negative values, the shading sequence is rescaled to the range $[-1,1]$ (by applying a linear transformation $\mathbf{S} \mapsto \mathbf{S} \times 2 - 1$) to match the input requirements of the VAE encoder~\citep{blattmann2023stable}. This rescaling makes the shading sequence more sensitive to geometric variations. The effectiveness of this strategy is validated in~\Tref{tab:abla} with comparison between `ours' and `ours w/o clamp'.

\begin{table}[t]
\caption{Quantitative analysis in terms of MAE and SNE of the normal map on LUCES benchmark dataset~\citep{mecca2021luces}. Highlighted numbers indicate the \colorbox{LightGreen!40}{best} and \colorbox{LightSkyBlue!40}{second best} results.}
\label{tab:abl_detail}
\footnotesize
\setlength{\tabcolsep}{2mm}
\resizebox{\linewidth}{!}{%
\begin{NiceTabular}{r|cccccccc}
\toprule
& GeoWizard & DSINE & StableNormal & Lotus-D & Lotus-G & Neural LightRig & NiRNE & \textbf{Ours} \\
\midrule
MAE ($\downarrow$) & 22.49 & 21.82 & 20.34 & 18.56 & \cellcolor{LightSkyBlue!40}17.44 & 20.95 & 17.88 & \cellcolor{LightGreen!40} \textbf{14.48} \\
SNE ($\downarrow$) & 37.76 & 33.08 & 29.20 & 33.20 & 29.85 & 32.77 & \cellcolor{LightSkyBlue!40}26.78 & \cellcolor{LightGreen!40} \textbf{26.74} \\
\bottomrule
\end{NiceTabular}
}
\vspace{-10pt}
\end{table}

\begin{table*}[t]
    \caption{Ablation study in terms of MAE of the normal map on LUCES benchmark dataset~\citep{mecca2021luces}. In particular, ``+M''(``+L'') means training on Multishade (LightProp) dataset, `w/o clamp' means removing clamping on shading sequence. `w/o MA' means training on dataset without material augmentation. Highlighted numbers indicate the \colorbox{LightGreen!40}{best} and \colorbox{LightSkyBlue!40}{second best} results.}
    \centering
    \label{tab:abla}
    \resizebox{\textwidth}{!}{
    \begin{NiceTabular}{l|cccccccccccccc|c}
        \toprule
        Method & {\sc Ball} & {\sc Bell} & {\sc Bowl} & {\sc Buddha} & {\sc Bunny} & {\sc Cup} & {\sc Die} & {\sc Hippo} & {\sc House} & {\sc Jar} & {\sc Owl} & {\sc Queen} & {\sc Squirrel} & {\sc Tool} & Mean \\
        \midrule
        Lotus-G & 17.82 & 8.66 & 10.89 & 19.71 &  12.90 & 23.26 & 12.59 & 16.94 & 35.32 & 10.69 & 18.94 & 20.65 & 24.05 &  11.74 &  17.44 \\
        Neural LightRig & 9.52 & 11.95 & 21.71 & 20.66 & 15.25 & 18.08 & 25.13 & 18.54 & 39.67 & 19.78 & 23.40 & 23.35 & 25.32 & 20.97 & 20.95 \\
        \midrule
        Ours w/o clamp & 11.63 & 7.75 & 12.66 & 17.79 & 12.32 & 15.72 & 9.30 & 14.08 & 40.77 & 4.64 & 20.65 & 21.18 & 21.01 & 12.84 &  15.88\\
        Ours w/o MA & 12.97 & \cellcolor{LightSkyBlue!40}6.92 & 9.25 & 19.00 & 14.90 &  15.96 & 10.65 & 15.18 & 39.35 & 6.64 & 20.49 & 20.65 & 21.50 & 13.23 &  16.19\\
        Lotus-G+M & 16.21 & 8.95 & \cellcolor{LightSkyBlue!40}7.11 & \cellcolor{LightGreen!40}16.57 & \cellcolor{LightGreen!40}11.50 & 22.41 & 16.40 & \cellcolor{LightSkyBlue!40}14.04 & 38.90 & 13.14 & 24.69 & \cellcolor{LightSkyBlue!40}19.78 & \cellcolor{LightGreen!40}18.96 & 14.35 & 17.36 \\
        Ours+L & 9.37 & 7.03 & 8.46 & 19.42 & \cellcolor{LightSkyBlue!40}12.24 & 14.05 & 8.73 & 15.06 & 38.14 & \cellcolor{LightSkyBlue!40}5.81 & 20.94 & 20.22 & 22.19 & \cellcolor{LightSkyBlue!40}11.23 & 15.21 \\

        Ours w/ spiral & 16.32 & 9.25 & 10.97 & 20.23 & 16.78 & 16.70 & 12.17 & 16.06 & 39.68 & 7.38 & 22.25 & 21.86 & 22.50 & 14.29 & 17.60 \\
        Ours w/ RGB & 10.36 & 8.56 & 8.99 & 18.28 & 13.89 & \cellcolor{LightGreen!40}11.56 & 9.22 & \cellcolor{LightGreen!40} 13.70 & 38.49 & \cellcolor{LightGreen!40}5.74 & 19.80 & 20.45 & 20.78 & 13.99 & 15.27 \\
        Ours w/ SVD XL & \cellcolor{LightGreen!40}8.69 & 7.68 & 9.16 & 18.34 & 12.43 & \cellcolor{LightSkyBlue!40}12.15 & \cellcolor{LightSkyBlue!40} 8.47 & 14.11 & \cellcolor{LightSkyBlue!40} 37.60 & 6.94 & \cellcolor{LightGreen!40}18.73 & \cellcolor{LightGreen!40}19.10 & \cellcolor{LightSkyBlue!40} 19.28 & 11.38 & \cellcolor{LightSkyBlue!40}14.58 \\
        \textbf{Ours} &  \cellcolor{LightSkyBlue!40}\textbf{9.09} &  \cellcolor{LightGreen!40}\textbf{5.94} &  \cellcolor{LightGreen!40}\textbf{6.84} &  \cellcolor{LightSkyBlue!40}\textbf{17.58} &  \textbf{12.70} &  \textbf{13.80} &  \cellcolor{LightGreen!40}\textbf{8.26} &  \textbf{14.14} & \cellcolor{LightGreen!40}\textbf{36.79} &  \textbf{5.93} &  \cellcolor{LightSkyBlue!40}\textbf{19.60} &  \textbf{19.99} &  \textbf{21.34} &  \cellcolor{LightGreen!40}\textbf{10.66} &  \cellcolor{LightGreen!40}\textbf{14.48}\\
        \bottomrule
    \end{NiceTabular}    
    }
    \vspace{-10pt}
\end{table*}

\textbf{Validation on material augmentation}. To evaluate the effectiveness of the proposed dataset, we train \ours~on the publicly available LightProp~\citep{he2024neural} dataset as a reference. The key difference is that our dataset introduces additional material augmentation to increase material diversity. As shown in~\Tref{tab:abla}, training without material augmentation (w/o MA, using only the original object materials) yields slightly worse performance on the LUCES~\citep{mecca2021luces} benchmark, while enabling material augmentation leads to notable improvements. These results demonstrate the effectiveness of the material augmentation.

\textbf{Validation on dataset impact}. We retrain the previous SOTA method on LUCES (\ie, Lotus-G) using our dataset (``Lotus-G+M''), resulting in consistent improvements and further validating the effectiveness of the proposed dataset. More importantly, when trained on the same dataset, our method still outperforms the baselines: Neural LightRig vs. ``Ours+L'' and ``Lotus-G+M'' vs. Ours, clearly demonstrating SOTA performance and highlighting its efficiency and competitiveness.
Finally, we also observed that for some specific objects, such as {\sc House}, retraining Lotus-G with our dataset resulted in decreased performance (35.32$^\circ$ for Lotus-G vs. 38.90$^\circ$ for ``Lotus-G+M''). This suggests that dataset variations may affect estimation accuracy on certain objects.

\blue{\textbf{Validation on model variants impact}. We train RoSE using a different video diffusion backbone, namely Stable Video Diffusion XL (SVD XL)~\citep{blattmann2023stable}, on the MultiShade dataset. We denote this variant as ``Ours w/ SVD XL''. As shown in \Tref{tab:abla}, this model achieves performance comparable to RoSE built on SV3D (14.58$^\circ$ for ``Ours w/ SVD XL'' and 14.48$^\circ$ for Ours). This demonstrates that our framework generalises well even when the backbone is pretrained on large-scale, general-purpose video data rather than a domain-specific object-centric dataset.}

\blue{\textbf{Validation on gray-scale input}. We train a variant of RoSE that replaces the grayscale input with an RGB input (\ie, Ours w/ RGB). The performance drops by $0.79^\circ$ on the LUCES benchmark. The result indicates the effectiveness of the grayscale input, which eliminates redundant chromatic information to enable accurate shading sequence estimation. }

\blue{\textbf{Validation on ring-light setup}. We train RoSE using a different light path where the elevation decreases from $60^\circ$ to $30^\circ$ while rotating $360^\circ$ around the $z$-axis. We denote this variant as ``Ours w/ spiral''. As shown in \Tref{tab:abla}, this more complex light path leads to a performance drop (MAE of $17.60^\circ$). This result highlights that the proposed ring-light setup is an effective and efficient design for predicting the shading sequence.}

\section{Discussion}
\label{sec:conclusion}
\textbf{Conclusion}.
We propose \ours, a novel method for monocular normal estimation that addresses the limitations of previous methods in their training paradigms to reduce 3D misalignment. By reformulating normal estimation to shading sequence estimation, \ours~facilitates normal estimation through an image-to-video generative model and a simple analytical solver. To further improve performance across more general scenarios, we train \ours~on~\dataset, a large-scale dataset with diverse materials and lighting conditions. Experiments show that \ours~outperforms state-of-the-art methods.

\textbf{Limitations \& Future Work}.
While {\ours} demonstrates strong performance in normal estimation across various settings, it has several limitations. First, employing video diffusion models for shading sequence generation introduces additional computational overhead, which may limit the applicability of the method in real-time scenarios. Second, {\ours} may struggle under extreme lighting conditions, particularly when large regions of the object receive insufficient illumination, resulting in degraded shading quality and less reliable normal predictions in those areas. 
\blue{Third, RoSE fails to produce high-quality normal maps on transparent or semi-transparent objects, and extending support for such cases will be an important direction for future work.}
Finally, the current evaluation is primarily object-centric, with a focus on robustness to varying light sources and reflectance properties. Extending {\ours} to scene-centric settings remains an important direction for future work.

\textbf{Acknowledgment}. This work is supported by the Ministry of Education, Singapore, AcRF Tier 1 grant No. RG98/24, and ByteDance Inc.

\newpage
\bibliography{iclr2026_conference}
\bibliographystyle{iclr2026_conference}

\clearpage
\input{supp.tex}

\end{document}

%% file: math_commands.tex
\usepackage{amsmath,amsfonts,bm}

\def\eqref#1{equation~\ref{#1}}

\def\1{\bm{1}}

\DeclareMathAlphabet{\mathsfit}{\encodingdefault}{\sfdefault}{m}{sl}
\SetMathAlphabet{\mathsfit}{bold}{\encodingdefault}{\sfdefault}{bx}{n}



%% file: supp.tex
\appendix
\makeatother
\tableofcontents

\section{LLM Usage Statement}
Large Language Models (LLMs) were only used in this research for writing optimization and grammar checking. No part of the theoretical contributions, experimental design, data analysis, or results was generated by LLMs.

\section{Broader Impact} 
The proposed method improves the accuracy of normal map estimation from a monocular image, with broad benefits across various applications. More precise geometry understanding can significantly improve downstream tasks such as 3D reconstruction, augmented reality, robotics, and digital content creation, enabling more immersive and interactive user experiences.

\section{Additional Implementation Details}
\textbf{Training details}.
As stated in the main paper, we build \ours~on top of SV3D~\citep{voleti2024sv3d}. During training, we initialize our model with \blue{pretrained SV3D weights (1.5B)} and fine-tune the first convolutional layer as well as all parameters within the self-attention and cross-attention modules \blue{on the MultiShade dataset, which contains 90K objects (3.1M image–normal pairs), and evaluated on a validation set of 100 objects, each providing 100 image–normal pairs}, \blue{result in 200M trainable parameters.} Training is conducted for 80,000 steps with a learning rate of $1\times10^{-5}$ and a total batch size of 16. We use the AdamW optimizer with $\beta_1 = 0.9$, $\beta_2 = 0.999$, and $1\times10^{-8}$ for weight decay. Training is performed in float16 precision for efficiency, and we apply gradient clipping with a maximum norm of 1.0. The model is trained to predict a 9-frame shading sequence, with each frame at a resolution of $576 \times 576$. End-to-end training takes approximately one day on 8 NVIDIA H100 80GB GPUs.

\textbf{Testing details}.
We follow the requirements specified in the baselines’ inference code~\citep{he2024neural,fu2024geowizard,ye2024stablenormal,ye2025hi3dgen,bae2024dsine} to prepare our test dataset, ensuring compatibility with each setup for a fair comparison. For both the LUCES~\citep{mecca2021luces} and DiLiGenT~\citep{shi2016benchmark} datasets, we use images indexed from 21 to 30 for testing, as the lights are more centered on the objects. \blue{All testing processes are performed on a single RTX A6000 Ada GPU. The total runtime includes the video diffusion model inference with 25 denoising steps and the shading to normal computation, the latter adding only a negligible cost of 0.045 seconds per object. For completeness, we also report the inference time of other methods for reference.}
\begin{table}[h]
\centering
\caption{\blue{Average inference time of monocular normal estimation methods per image (in seconds).}}
\label{tab:time}
\resizebox{\textwidth}{!}{
\begin{tabular}{l|ccccccc|c}
\toprule
Method & GeoWizard & DSINE & StableNormal & Lotus G & Lotus D & Neural LightRig & NiRNE & Ours \\
\midrule
Time & 101.11 & 0.83 & 1.52 & 0.61 & 0.59 & 93.73 & 0.31 & 10.57 \\
\bottomrule
\end{tabular}
}
\end{table}

\section{Additional Details about~\dataset}
\textbf{More Details about Material Augmentation}.
We present a statistical comparison of the proposed dataset with other related datasets~\citep{he2024neural,ye2025hi3dgen,jin2025neural,ikehata2022universal,ikehata2023scalable} in~\Tref{tab:dataset_statistics}, including works~\citep{he2024neural,ye2025hi3dgen} that are either recently released or not yet publicly available. 
We apply material augmentation (MA) with a probability of 0.5 by randomly replacing an object’s material with one sampled from the MatSynth dataset~\citep{vecchio2024matsynth}, selecting equally from metallic (617) or non-metallic (5,040) material groups. This process yields an additional 42,732 objects that share the same 3D geometry but differ in material appearance. The resulting MultiShade dataset, enriched with material diversity and rendered shading sequences, enables our method to achieve state-of-the-art performance on public benchmarks.
\begin{table}[h]
\vspace{-10pt}
\setlength{\tabcolsep}{5mm}
\caption{Statistics of representative datasets used for normal estimation under arbitrary lighting. \#$O$ and \#$v$ denote the number of 3D models and rendered views, respectively. N.A. indicates that the corresponding information is not available. `env.', `par.', `poi.' stand for environment light, parallel lights, and point lights, respectively.}
    \centering
    \resizebox{\linewidth}{!}{
    \begin{NiceTabular}{l|cccc}
        \toprule
        \textbf{Dataset} & \textbf{\# $O$} & \textbf{\# $v$} &\textbf{Light Compose} & \multicolumn{1}{c}{\textbf{Material}} \\
        \midrule
        PS-Wild~\citep{ikehata2023scalable} & 410   & 1 & env.(31)~/par.~/poi. & AdobeStock (926) \\
        PS-Mix~\citep{ikehata2023scalable} & 480   & 1      & env.(31)~/par.~/poi. & AdobeStock (897) \\
        LightProp~\citep{he2024neural} & 80K   & 5  & env.(24)~/area~/poi. & Objaverse \\
        RelitObjaverse~\citep{jin2025neural} & 90K   & 16  & env.(1,870)~/area & Objaverse \\
        DetailVerse~\citep{ye2025hi3dgen} & 700K   & 40  & N.A. & N.A. \\
        \midrule
        Ours & \textbf{90K}  & \textbf{6}  & \textbf{env.(780)~/par.~/poi.} & \textbf{Objaverse + MatSynth (5,657)} \\
        \bottomrule
    \end{NiceTabular}%
    }
    \label{tab:dataset_statistics}
\end{table}

\begin{figure}[ht]
  \vspace{-10pt}
  \centering
  \includegraphics[width=\textwidth]{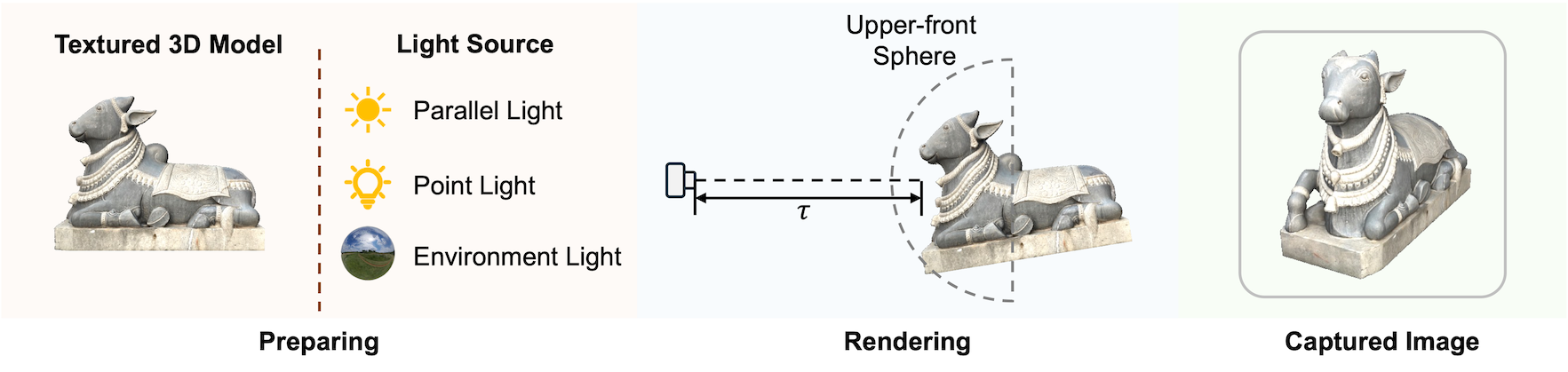}
  \caption{Image rendering setup.}
  \label{fig:setup}
\end{figure}

\textbf{Rendering setup}.
We construct our dataset using the Cycles rendering engine in Blender~\citep{blender}, selecting 90,546 filtered objects from Objaverse~\citep{jin2025neural}. Each object is rendered from six viewpoints. 
For each view, we implement one parallel light, one point light, or two HDR environment maps, selected from a pool of 780 real-world HDR environments~\citep{polyheaven}. The directions of the point and parallel lights are randomly sampled from the upper-front hemisphere facing the camera (see \Fref{fig:setup}). The camera is positioned at a random distance $\tau$ between 1.5 and 1.8 meters from the object, with a focal length of 35 mm, following the setup in~\citep{liu2023zero}.

\newpage
\clearpage
\section{Results on Web Images}
\begin{figure}[ht]
    \centering
    \includegraphics[width=0.77\linewidth]{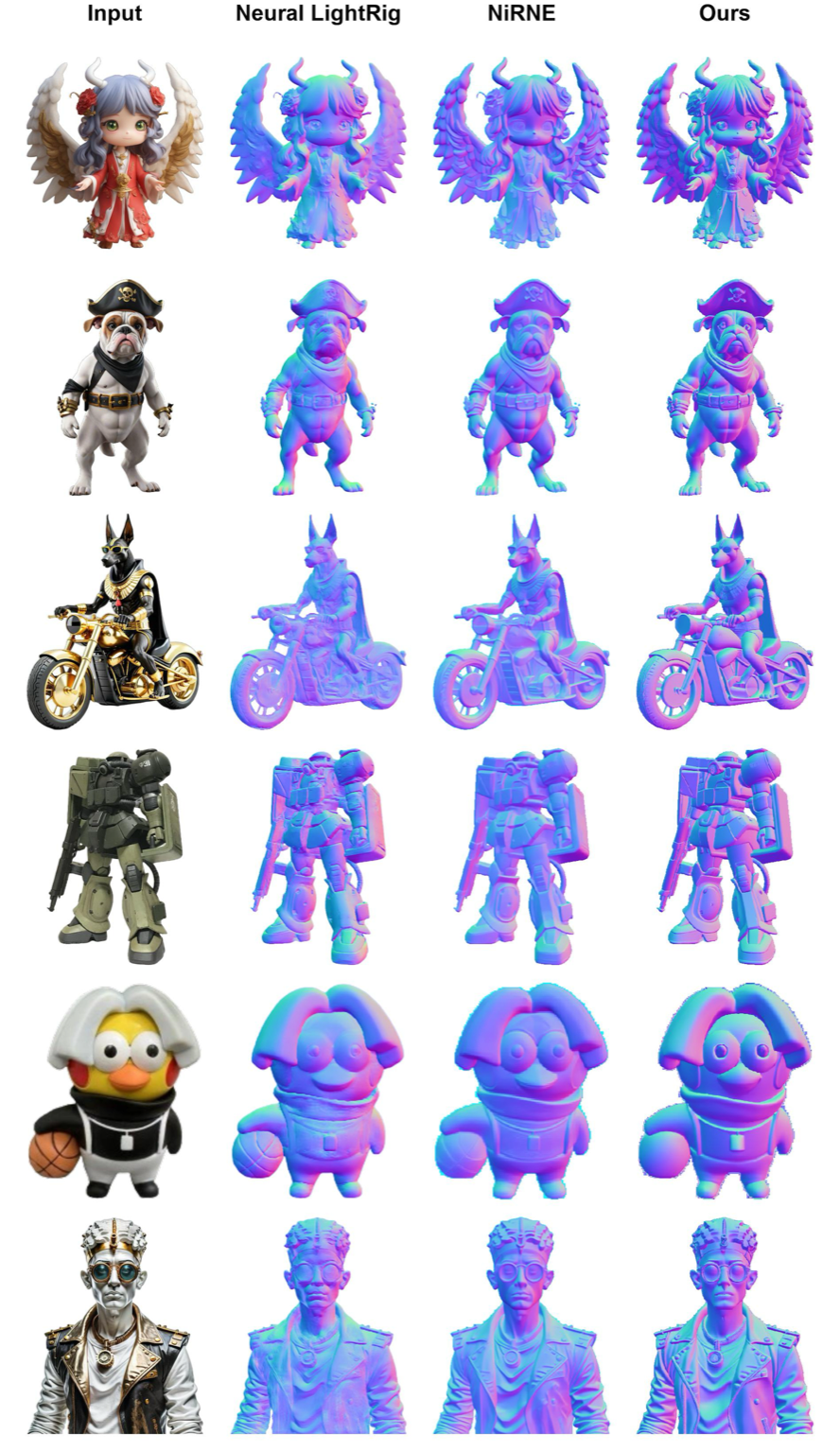}
    \caption{Qualitative comparison of normal maps on web images.}
    \label{fig:webimage_1}
\end{figure}

\begin{figure}[h]
    \centering
    \includegraphics[width=0.92\linewidth]{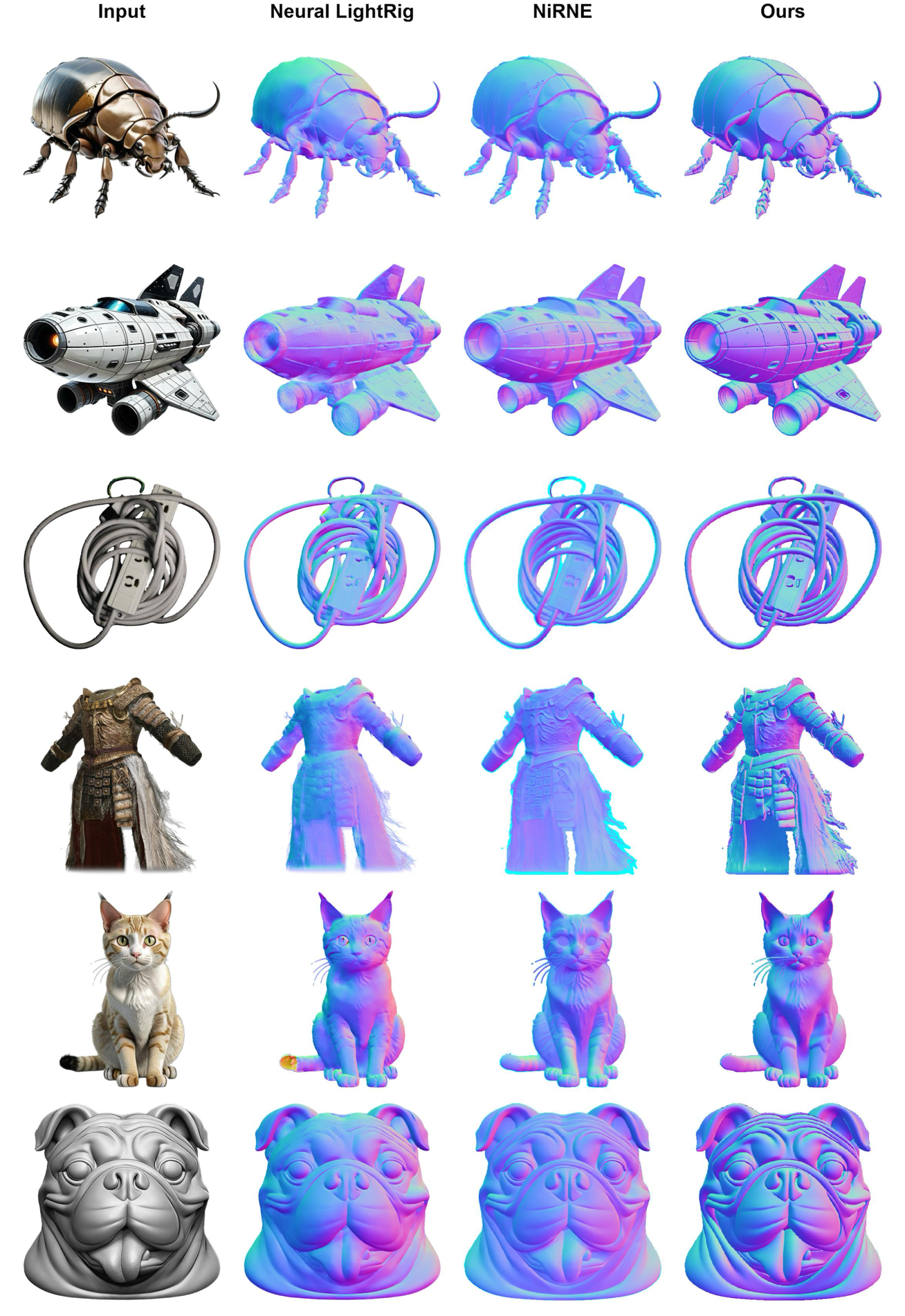}
    \caption{Qualitative comparison of normal maps on web images.}
    \label{fig:webimage_2}
\end{figure}

We present qualitative comparisons with state-of-the-art methods (NiRNE~\citep{ye2025hi3dgen} and Neural LightRig~\citep{he2024neural}) on additional images sourced from public resources, including the project page of StableNormal~\citep{ye2024stablenormal} and Google Images, as shown in \Fref{fig:webimage_1} and \Fref{fig:webimage_2}. The surface reconstruction from normals is performed using the method from~\citep{cao2022bilateral}.


\clearpage
\newpage
\section{Additional Experiment Results on Popular Datasets}
\subsection{Results on DiLiGenT}
We present a qualitative comparison of different methods for normal estimation. To avoid excessive redundancy, we select the normal map whose MAE is closest to the average MAE as a representative example for reference.

\begin{figure}[ht]
    \centering
    \includegraphics[width=\linewidth]{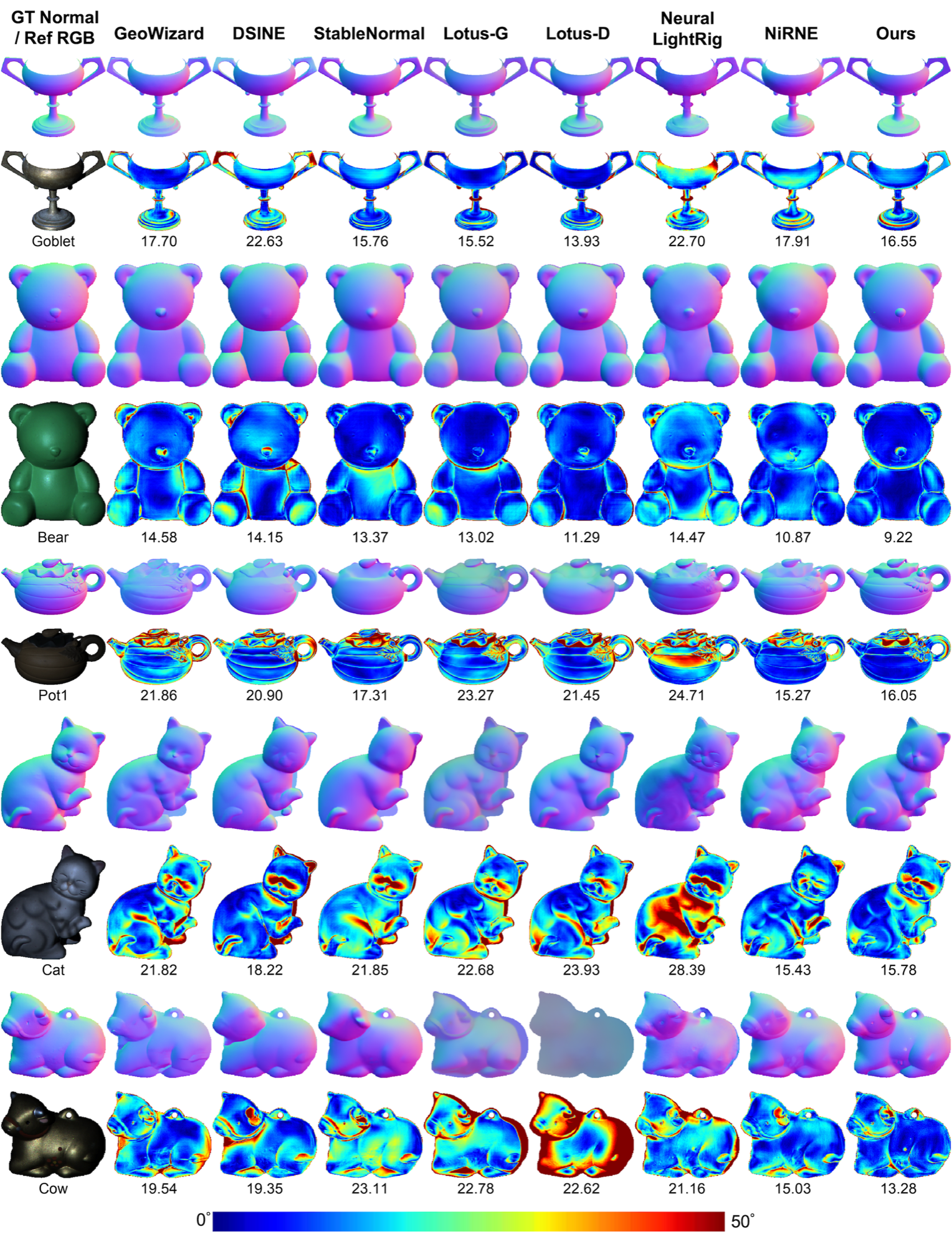}
    \caption{Qualitative comparison on normal maps and error maps for the {\sc Goblet}, {\sc Bear}, {\sc Pot1}, {\sc Cat}, {\sc Cow} from the DiLiGenT~\citep{shi2016benchmark} benchmark.}
    \label{fig:diligent1}
\end{figure}

\begin{figure}[h]
    \centering
    \includegraphics[width=\linewidth]{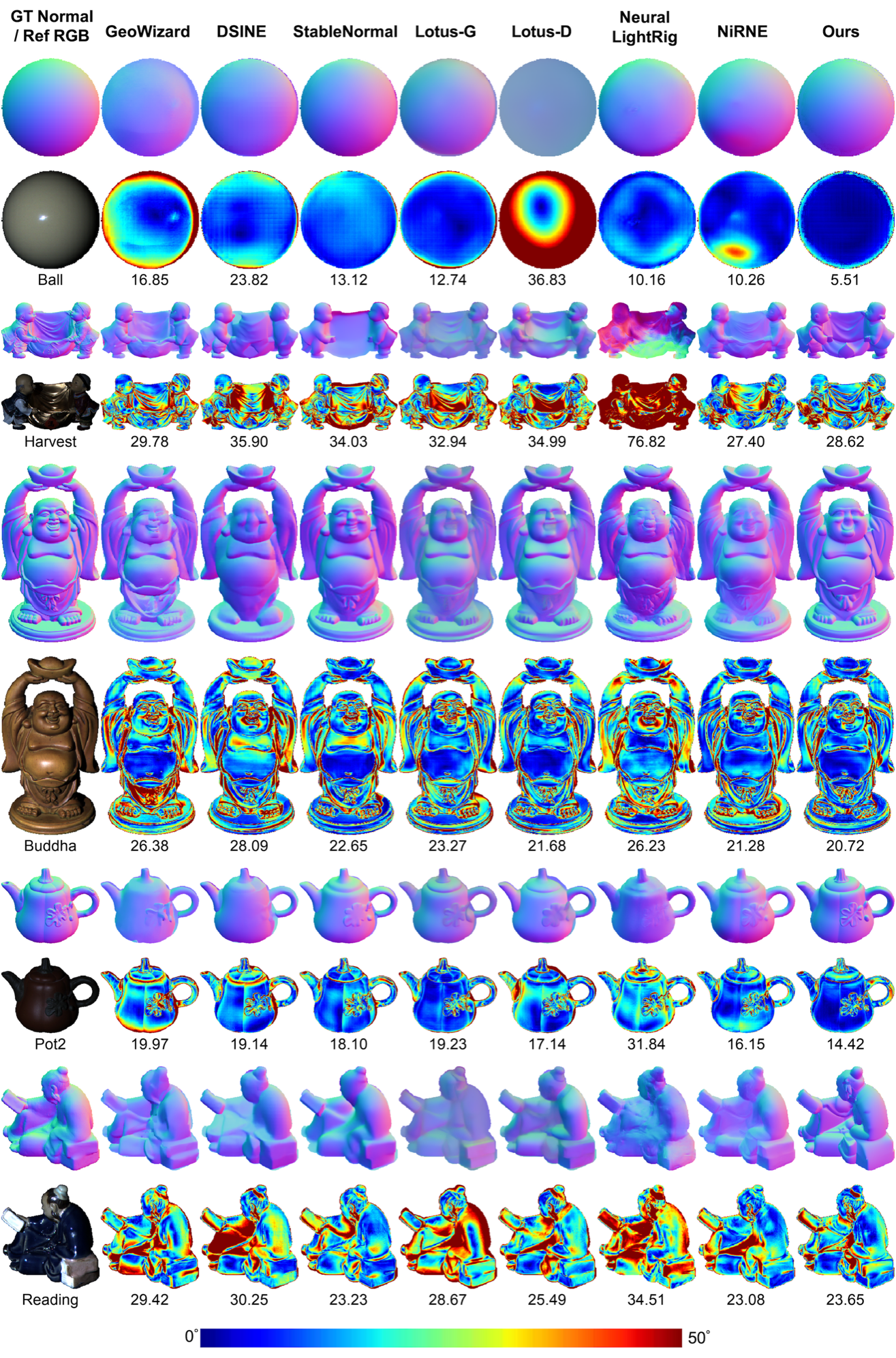}
    \caption{Qualitative comparison on normal maps and error maps for the {\sc Ball}, {\sc Harvest}, {\sc Buddha}, {\sc Pot2}, {\sc Reading} from the DiLiGenT~\citep{shi2016benchmark} benchmark.}
    \label{fig:diligent2}
\end{figure}
\clearpage
\newpage
\subsection{Results on LUCES}
We present a qualitative comparison on LUCES~\citep{mecca2021luces} of different methods for normal estimation. To avoid excessive redundancy, we select the normal map whose MAE is closest to the average MAE as a representative example for reference.

\begin{figure}[h]
    \centering
    \includegraphics[width=\linewidth]{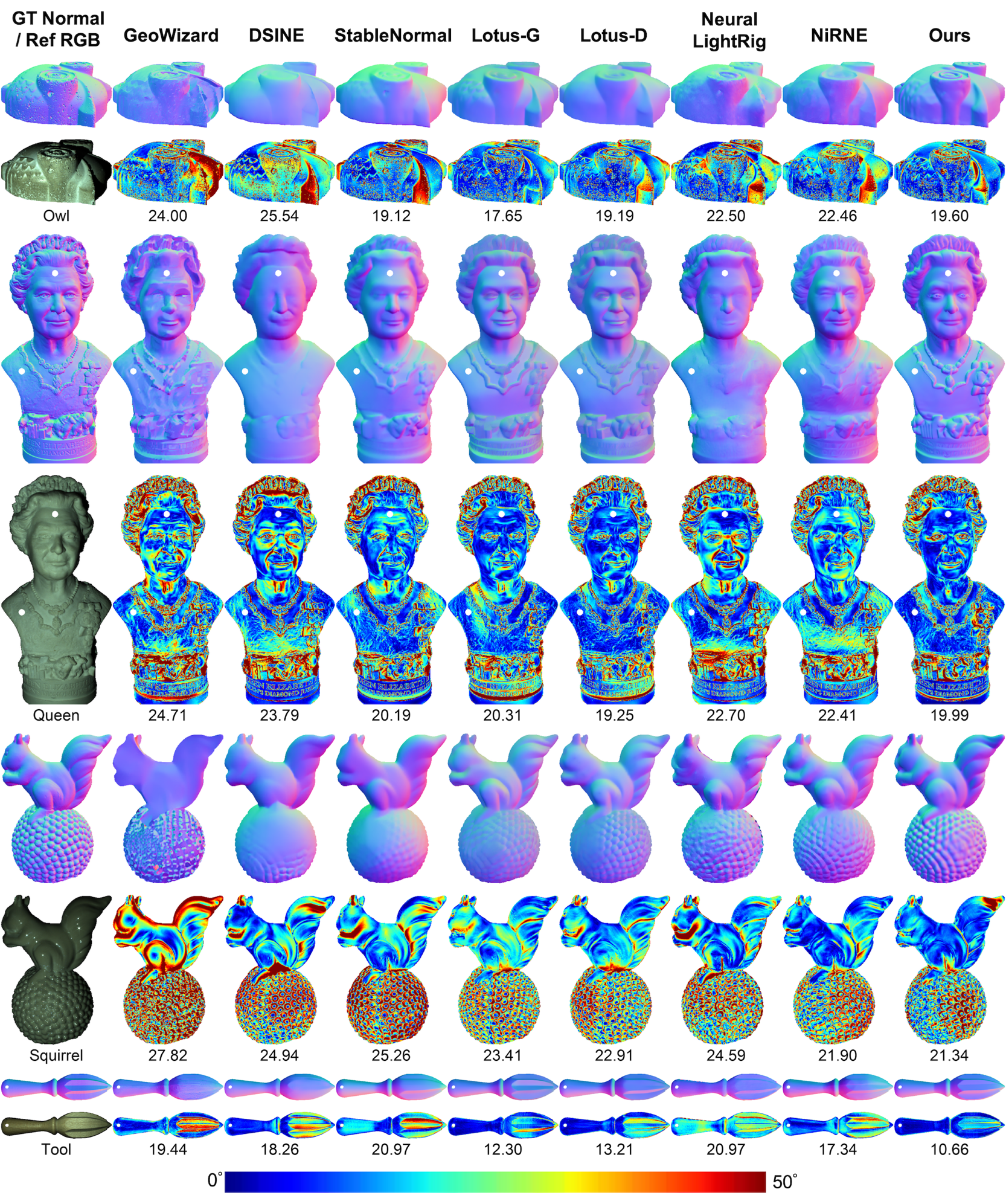}
    \caption{Qualitative comparison on normal maps and error maps for the {\sc Owl}, {\sc Queen}, {\sc Squirrel}, {\sc Tool} from the LUCES~\citep{mecca2021luces} benchmark.}
    \label{fig:luces-1}
\end{figure}

\begin{figure}[h]
    \centering
    \includegraphics[width=\linewidth]{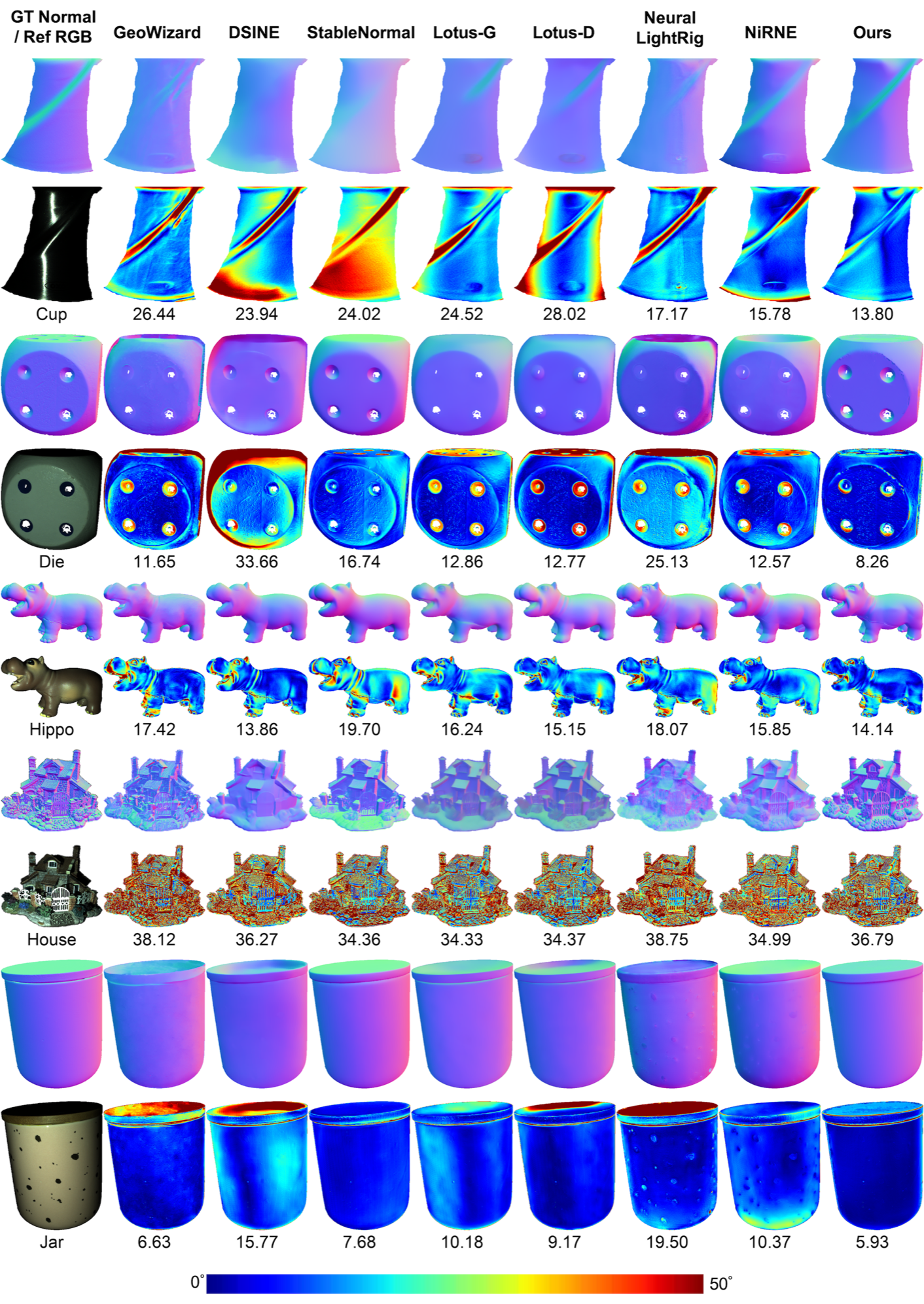}
    \caption{Qualitative comparison on normal maps and error maps for the {\sc Cup}, {\sc Die}, {\sc Hippo}, {\sc House}, and {\sc Jar} from the LUCES~\citep{mecca2021luces} benchmark.}
    \label{fig:luces-2}
\end{figure}

\begin{figure}[h]
    \centering
    \includegraphics[width=\linewidth]{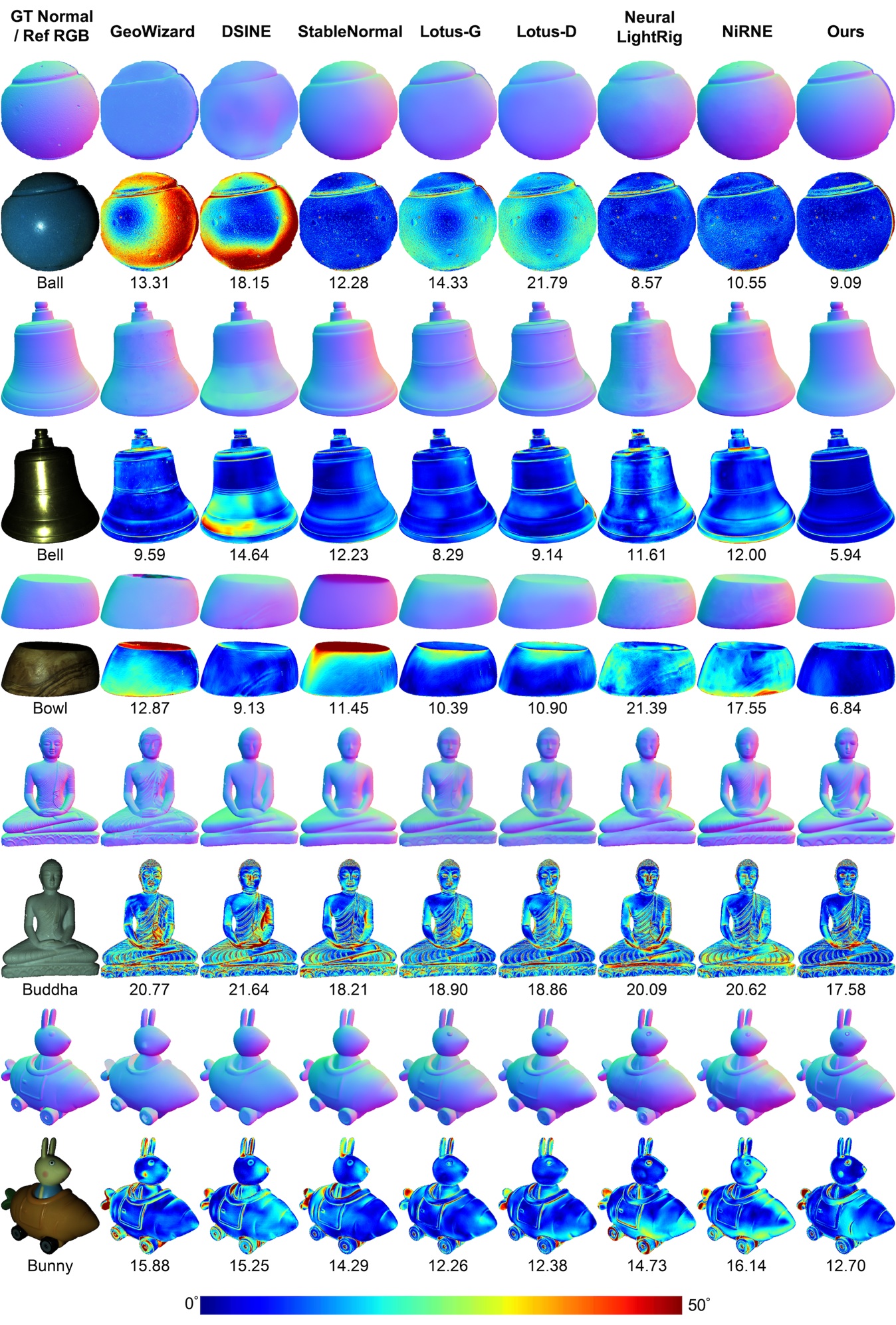}
    \caption{Qualitative comparison on normal maps and error maps for the {\sc Owl}, {\sc Queen}, {\sc Squirrel}, {\sc Tool} from the LUCES~\citep{mecca2021luces} benchmark.}
    \label{fig:luces-3}
\end{figure}

\clearpage
\newpage

\subsection{Results on LightProp}
We report quantitative results on the LightProp dataset~\citep{he2024neural}. Our method achieves the second-best overall performance in terms of mean and median angular errors for normal map estimation. We re-ran the entire evaluation, as we observed discrepancies between our results and those reported in the original paper~\citep{he2024neural}. The metric computation scripts used in our evaluation follow the implementation in~\citep{bae2024dsine}\footnote{\url{https://github.com/baegwangbin/DSINE/blob/main/projects/dsine/test.py}}.
\begin{table*}[ht]
\setlength{\tabcolsep}{3mm}
    \centering
    \caption{Quantitative comparison in terms of \textbf{Mean} and \textbf{Median} Angular Errors of the normal map on LightProp test set, and the percentage of objects below a specific error bound. \textbf{Bold} (\underline{underline}) numbers indicate the \textbf{best} (\underline{second-best}) results among single-view normal estimation methods.}
    \label{tab:lightprop}
    \footnotesize
    \resizebox{\textwidth}{!}{
    \begin{NiceTabular}{l|cc|cccccc}
        \toprule
        Method & Mean $\downarrow$ & Median $\downarrow$ & $3^{\circ}$(\%) $\uparrow$& $5^{\circ}$(\%) $\uparrow$& $7.5^{\circ}$(\%) $\uparrow$& $11.25^{\circ}$(\%) $\uparrow$& $22.5^{\circ}$(\%) $\uparrow$& $30^{\circ}$(\%) $\uparrow$\\
        \midrule
                GeoWizard & 21.03 & 13.07 & 9.94  & 20.29 & 31.63 & 44.87 & 68.23 & 76.97 \\
    DSINE & 22.16 & 14.02 & 9.48  & 18.00 & 28.31 & 41.85 & 66.99 & 76.12 \\
    StableNormal & 19.66 & 12.98 & 3.78  & 10.64 & 23.15 & 42.50 & 73.85 & 82.37 \\
    Lotus-D & 19.10 & 12.26 & 10.26 & 20.97 & 32.51 & 46.83 & 71.83 & 80.45 \\
    Lotus-G & 19.19 & 12.15 & 10.91 & 22.06 & 33.91 & 47.35 & 71.12 & 79.70 \\
    Neural LightRig & \textbf{15.29} & \textbf{8.84} & \textbf{20.13} & \textbf{32.33} & \textbf{44.68} & \textbf{57.99} & \textbf{78.99} & \textbf{85.79} \\
    NiRNE & 17.87 & 12.38 & 7.21  & 16.69 & 29.13 & 45.68 & 75.01 & \underline{84.02} \\
    \midrule
    Ours  & \underline{17.40} & \underline{11.00} & \underline{17.15} & \underline{26.49} & \underline{37.33} & \underline{50.79} & \underline{75.29} & 83.10 \\
    \bottomrule
    \end{NiceTabular}
}
    \vspace{-0.2in}

\end{table*}

\newpage
\subsection{Results on Natural Light Photometric Stereo Dataset}
We conduct quantitative comparison results on a synthetic dataset (NaPS) proposed in~\citep{li2024spin}, which contains a series of rendered images based on objects selected from~\citep{shi2016benchmark}, under varying environment lighting, material properties, and object shapes. The dataset is organized into four groups: (1) the light group, which evaluates performance on a single object under different lighting conditions; (2) the shape group, which compares different object geometries; (3) the reflectance group, which assesses performance on diffuse and specular materials; and (4) the spatially varying material group, which poses a challenging scenario with complex, spatially varying materials. During testing, we select the 10 images with the highest average brightness for evaluation. Our method achieves either the best or second-best performance in each group and ranks first in terms of overall average performance ($11.35^\circ$, ours vs. $12.32^\circ$, NiRNE, the second best method).
\begin{table}[ht]
\setlength{\tabcolsep}{1.5mm}
  \centering
  \caption{Results on the NaPS~\citep{li2024spin} dataset for normal estimation under natural lighting. `L., A., S., U.' denote four types of environment maps, Landscape, Attic, Studio, and Urban, covering both outdoor and indoor settings. `D., S.' in the reflectance and spatially varying material groups refer to diffuse and specular materials, respectively.  Numbers indicate the MAE ($\downarrow$) of the estimated normal maps. \textbf{Bold} (\underline{underline}) numbers indicate the \textbf{best} (\underline{second-best}) results.}
  \resizebox{\linewidth}{!}{
         \begin{NiceTabular}{l|ccccc|ccccc}
    \toprule
    \multirow{2}[4]{*}{Method} & \multicolumn{5}{c|}{Light Group}      & \multicolumn{5}{c}{Reflectance Group} \\
\cmidrule{2-11}          & Cow (L.) & Cow (A.) & Cow (S.) & Cow (U.) & AVG   & Pot2 (D.) & Pot2 (S.) & Reading (D.) & Reading (S.) & AVG \\
    \midrule
    GeoWizard & 10.24 & 10.59 & 12.24 & 11.52 & 11.15 & 12.11 & 11.43 & 16.09 & 14.18 & 13.45 \\
    DSINE & 14.41 & 15.65 & 14.41 & 13.43 & 14.48 & 15.86 & 14.63 & 17.03 & 17.23 & 16.19 \\
    StableNormal & 14.58 & 12.49 & 16.11 & 16.61 & 14.95 & 10.44 & 11.24 & 14.87 & 13.54 & 12.52 \\
    Lotus-D & \textbf{8.43} & \textbf{9.34} & \underline{10.82} & \textbf{10.00} & \textbf{9.65} & \underline{10.00} & 10.03 & 14.53 & 13.14 & 11.92 \\
    Lotus-G & 11.66 & 11.28 & 12.48 & 10.29 & 11.43 & 12.00 & 11.51 & 15.93 & 13.67 & 13.28 \\
    Neural LightRig & \underline{9.25}  & \underline{9.90}  & 11.98 & 11.21 & 10.59 & 10.54 & 10.22 & 13.73 & 12.63 & 11.78 \\
    NiRNE & 9.89  & 10.66 & 12.72 & 10.55 & 10.95 & \textbf{7.66} & \textbf{8.66} & \textbf{11.21} & \textbf{10.81} & \textbf{9.58} \\
    Ours  & 9.85  & 10.06 & \textbf{10.51} & \underline{10.09} & \underline{10.13} & 10.74 & \underline{9.77}  & \underline{11.73} & \underline{11.07} & \underline{10.83} \\
    \midrule
    \multirow{2}[4]{*}{Method} & \multicolumn{5}{c|}{Shape Group}      & \multicolumn{5}{c}{Spatially Varying Material Group} \\
\cmidrule{2-11}          & Ball  & Bear  & Buddha & Reading & AVG   & Pot2 (D.) & Pot2 (S.) & Reading (D.) & Reading (S.) & AVG \\
    \midrule
    GeoWizard & \underline{3.93}  & 9.71  & 21.52 & 15.36 & 12.63 & 14.42 & 14.15 & 22.82 & 22.32 & 18.43 \\
    DSINE & 27.68 & 9.85  & 24.22 & 16.45 & 19.55 & 23.92 & 20.19 & 22.81 & 21.22 & 22.04 \\
    StableNormal & 8.33  & \textbf{8.04} & 16.77 & 13.64 & 11.69 & 15.71 & 15.10 & 19.77 & 18.90 & 17.37 \\
    Lotus-D & 8.71  & 9.30  & \textbf{16.15} & 13.50 & 11.91 & \underline{13.18} & \underline{13.54} & 21.95 & \underline{17.86} & \underline{16.63} \\
    Lotus-G & 11.18 & 9.51  & \underline{16.64} & 14.46 & 12.95 & 16.20 & 15.27 & 24.59 & 18.15 & 18.55 \\
    Neural LightRig & \textbf{3.11} & 9.18  & 17.37 & 13.88 & \underline{10.89} & 13.72 & 14.30 & 23.85 & 24.23 & 19.03 \\
    NiRNE & 8.57  & 9.45  & 18.37 & \underline{11.92} & 12.08 & 14.04 & 14.42 & \underline{19.11} & 19.10 & 16.67 \\
    Ours  & 5.25  & \underline{8.71}  & 17.56 & \textbf{11.23} & \textbf{10.69} & \textbf{11.98} & \textbf{11.87} & \textbf{16.06} & \textbf{15.05} & \textbf{13.74} \\
    \bottomrule
    \end{NiceTabular}%
    }
  \label{tab:addlabel}%
\end{table}%

\newpage
\clearpage
\subsection{Overall Performance Comparison}
In this paper, we conduct a comprehensive analysis across multiple benchmark datasets, including two real-world datasets (DiLiGenT~\citep{shi2016benchmark} and LUCES~\citep{mecca2021luces}) and three synthetic datasets (MultiShade, NaPS~\citep{li2024spin}, and LightProp~\citep{he2024neural}). According to~\Tref{tab:quantitative_all}, our method achieves the best average performance in terms of MAE and overall ranking. This demonstrates the strong generalization ability of the proposed~\ours~across diverse scenarios, lighting conditions, and object types.
\begin{table}[htbp]
  \centering
  \caption{Quantitative comparison over all five datasets. We report the MAE ($\downarrow$) over all objects and the rank ($\downarrow$) among methods at specific dataset for comparison.}
  \resizebox{\textwidth}{!}{
    \begin{NiceTabular}{lcc|cc|cc|cc|cc|cc}
    \toprule
    \multicolumn{1}{c}{\multirow{2}[4]{*}{Method}} & \multicolumn{2}{c}{DiLiGenT} & \multicolumn{2}{c}{LUCES} & \multicolumn{2}{c}{MultiShade} & \multicolumn{2}{c}{LightProp} & \multicolumn{2}{c}{NaPS} & \multicolumn{2}{c}{AVG} \\
\cmidrule{2-13}          & MAE($\downarrow$)   & Rank($\downarrow$)  & MAE($\downarrow$)   & Rank($\downarrow$)  & MAE($\downarrow$)   & Rank($\downarrow$)  & MAE($\downarrow$)   & Rank($\downarrow$)  & MAE($\downarrow$)   & Rank($\downarrow$)  & MAE($\downarrow$)   & Rank($\downarrow$) \\
    \midrule
    GeoWizard & 21.79 & 5.00  & 22.49 & 8.00  & 20.46 & 6.00  & 21.03 & 7.00  & 13.91 & 5.00  & 19.94 & 6.20 \\
    DSINE & 23.25 & 7.00  & 21.82 & 7.00  & 22.53 & 8.00  & 22.16 & 8.00  & 18.06 & 8.00  & 21.56 & 7.60 \\
    StableNormal & 20.44 & 3.00  & 20.34 & 5.00  & 19.71 & 5.00  & 19.66 & 6.00  & 14.13 & 7.00  & 18.86 & 5.20 \\
    Lotus-D & 22.94 & 6.00  & 18.56 & 4.00  & \underline{18.48} & \underline{2.00}  & 19.10 & 4.00  & 12.53 & 3.00  & 18.32 & 3.80 \\
    Lotus-G & 21.41 & 4.00  & \underline{17.44} & \underline{2.00}  & 18.76 & 3.00  & 19.19 & 5.00  & 14.05 & 6.00  & 18.10 & 4.00 \\
    Neural LightRig & 29.10 & 8.00  & 20.95 & 6.00  & 20.59 & 7.00  & \textbf{15.29} & \textbf{1.00} & 13.07 & 4.00  & 19.80 & 5.20 \\
    NiRNE & \underline{17.27} & \underline{2.00}  & 17.88 & 3.00  & 19.57 & 4.00  & 17.87 & 3.00  & \underline{12.32} & \underline{2.00}  & \underline{16.98} & \underline{2.80} \\
    \midrule
    Ours  & \textbf{16.36} & \textbf{1.00} & \textbf{14.48} & \textbf{1.00} & \textbf{15.37} & \textbf{1.00} & \underline{17.40} & \underline{2.00}  & \textbf{11.35} & \textbf{1.00} & \textbf{14.99} & \textbf{1.20} \\
    \bottomrule
    \end{NiceTabular}%
    }
  \label{tab:quantitative_all}%
\end{table}%

\clearpage
\newpage

\newpage
\clearpage
\section{\blue{Case Analysis}}
\begin{figure}[ht]
    \centering
    \includegraphics[width=\linewidth]{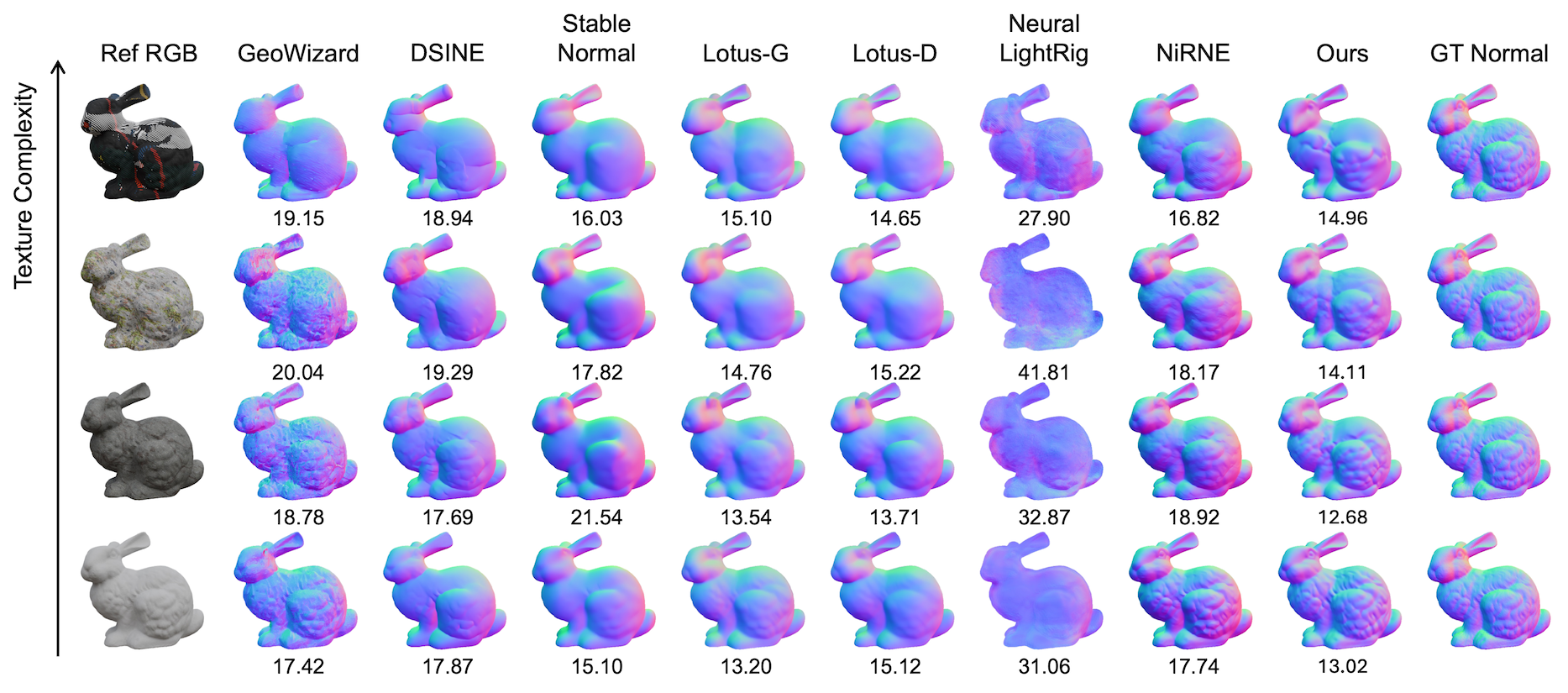}
    \caption{Analysis of each method’s performance across different texture settings on {\sc Bunny}.}
    \label{fig:case_01}
\end{figure}

\begin{figure}[ht]
    \centering
    \includegraphics[width=\linewidth]{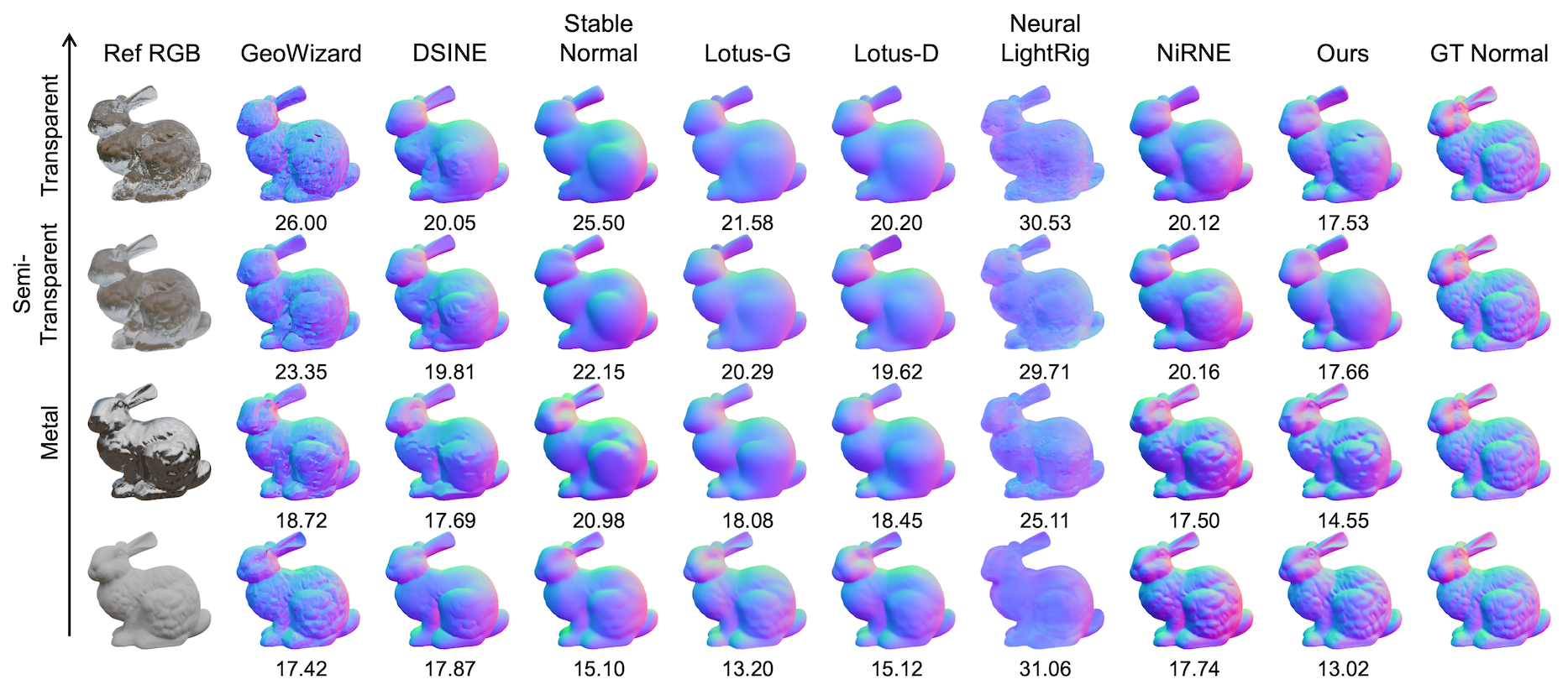}
    \caption{Analysis of each method’s performance across different BRDF settings on {\sc Bunny}.}
    \label{fig:case_02}
\end{figure}

\begin{figure}[ht]
    \centering
    \includegraphics[width=\linewidth]{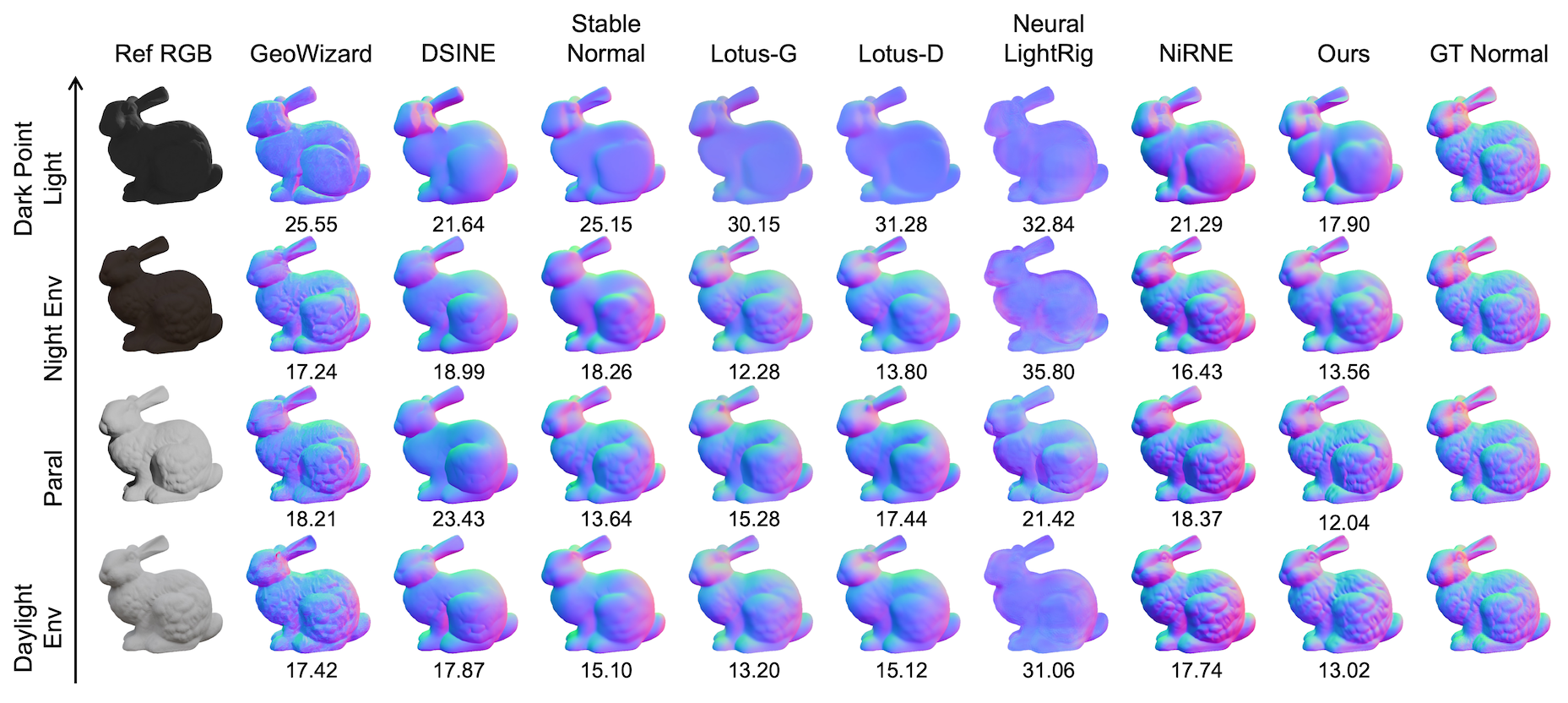}
    \caption{Analysis of each method’s performance across different light settings on {\sc Bunny}, where ``Paral'' indicate parallel lights, ``Env'' stands for environment light.}
    \label{fig:case_03}
\end{figure}

\blue{We conduct a detailed analysis across variations in lighting, textures, and BRDFs through three groups of controlled experiments. 
The first group examines how different lighting conditions, including simple light, parallel light, point light, and challenging light, influence performance, with texture and BRDF fixed. The second group evaluates the impact of texture complexity by comparing a simple texture with three spatially varying textures, while holding lighting and BRDF constant. The third group isolates material effects by varying the BRDF under simple lighting and texture settings. The results (see \Fref{fig:case_01}-\Fref{fig:case_03}) show that our method consistently produces high-quality normal maps relative to other approaches. However, performance does degrade under highly complex textures, challenging lighting conditions, or difficult material properties.}

\newpage
\clearpage
\blue{\section{Additional Discussion}}
\blue{\subsection{Analysis on Latent's Distribution}}

\begin{figure}[h]
    \centering    \includegraphics[width=\linewidth]{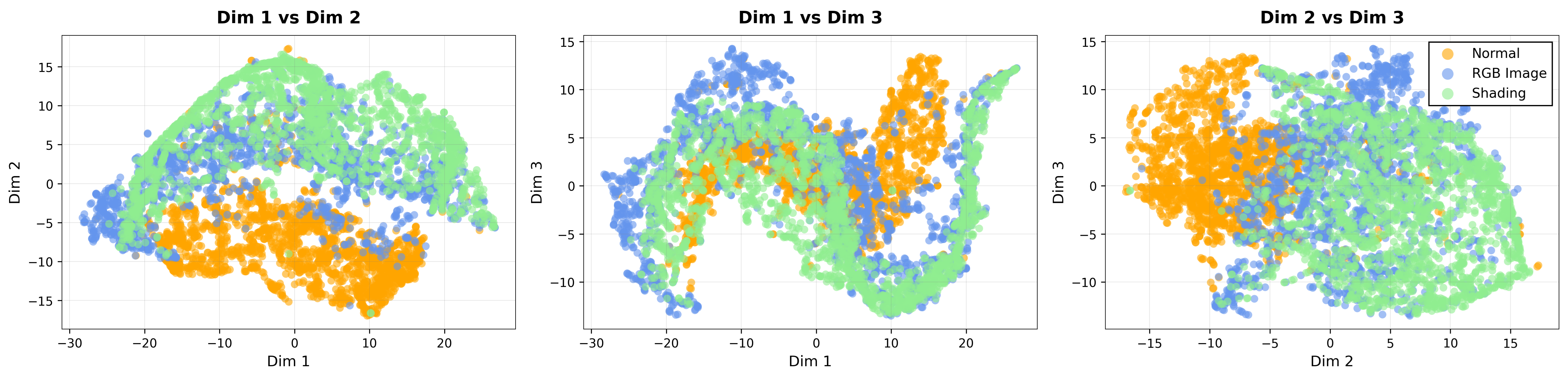}
    \caption{\blue{Visualization of the distribution of the downsampled latents of normal maps, RGB images, and shading maps from \dataset, shown from three orthogonally projected views.}}
    \label{fig:latent_tsne}
\end{figure}

\blue{We analyse the latent representations of RGB images, corresponding normal maps, and shading maps from the~\dataset~after encoding them with the Stable Diffusion~\citep{rombach2022high} VAE, in order to examine whether shading maps lie closer to RGB images than the normal map in latent space. Specifically, we randomly sample 1000 objects from the dataset. For each object, we randomly select one RGB image under a random viewpoint and lighting condition, retrieve its corresponding normal map, and randomly choose one shading map computed under 9 ring lights setup, giving us 3000 samples in total. Each sample has size 72 $\times$ 72 $\times$ 8. We average each latent across the 72 $\times$ 72 spatial dimensions and apply t-SNE to obtain a 3D embedding for each sample. This results in three sets of 1000 $\times$ 3 vectors, which we visualize in \Fref{fig:latent_tsne}. As shown, the shading latents exhibit a tendency to cluster closer to RGB image latents.}

\blue{\subsection{Discussion on Shading Sequence’s Robustness}
We conduct a 5-run experiment to evaluate how perturbations in the shading sequence affect the final normal estimation. In each run, identical Gaussian noise of varying magnitudes is added to one or multiple shading maps and the surface normal. The results on {\sc Bunny} from LUCES~\citep{mecca2021luces} show that the shading sequence is noticeably robust to the perturbations. When noise is injected into a single shading map, the change in mean angular error over 5 runs remains small ($1.537^\circ$ to $5.233^\circ$), and is substantially lower than perturbing the normal map directly ($3.235^\circ$ to $24.421^\circ$), even under the strongest disturbance. A larger deviation emerges only when 9 shading maps are perturbed simultaneously ($3.612^\circ$ to $20.713^\circ$), yet the shading-sequence formulation still remains more robust than directly perturbing normals. These observations confirm that predicting normals from shading sequences provides an inherent degree of robustness to noise.}

\begin{table}[h]
\centering
\caption{\blue{Quantitative analysis of how perturbations to shading sequences and normal maps affect the mean angular error relative to the clean version. `x shading maps' indicates that x frames in the shading sequence are perturbed by Gaussian noise. The first row specifies the noise magnitude.}}
\label{tab:noise_effect}
\setlength{\tabcolsep}{8mm}
\resizebox{\textwidth}{!}{
\begin{tabular}{l|ccccc}
\toprule
Perturbed Frames & 0.050 & 0.100 & 0.200 & 0.300 & 0.400 \\
\midrule
1 Shading Map   & 1.537 & 2.405 & 3.479 & 4.452 & 5.233 \\
5 Shading Maps  & 2.907 & 4.766 & 7.514 & 10.868 & 14.033 \\
9 Shading Maps  & 3.612 & 5.745 & 10.468 & 15.727 & 20.713 \\
Normal Map      & 3.235 & 6.454 & 12.764 & 18.825 & 24.421 \\
\bottomrule
\end{tabular}
}
\end{table}

\blue{\subsection{Discussion on Different Shading-based Paradigm}
To compare the two paradigms, \ie, (1) first estimating normals and then computing a shading loss, and (2) our proposed paradigm that directly predicts the shading sequence and then derives normals, we evaluate the effect of incorporating a rendering-based shading loss into an image diffusion model. Specifically, we apply the shading loss to Lotus-G~\citep{he2024lotus}, where the loss is defined as the L1 distance between the predicted shading sequence and the ground truth sequence. Each shading map is computed as the dot product between the predicted surface normal and the preset ring-light path, with negative values clamped to 0. This variant (``Lotus w/ shad.'') is trained on the MultiShade dataset.
The results show that Lotus w/ shad achieves improved performance (16.20$^\circ$ MAE on the LUCES benchmark). However, it still falls short of our method (14.48$^\circ$). This supports the effectiveness of our framework: estimating the entire shading sequence with a video diffusion model allows the network to capture the interrelationships among shading maps that contain rich geometric information, thereby improving the quality of the estimated shading sequence and enabling more accurate normal estimation.
}
\begin{table}[ht]
\centering
\caption{Comparison between ``Lotus w/ shad.'' and the proposed \ours~on the LUCES benchmark. Numbers indicate MAE ($\downarrow$) in degrees. \textbf{Bold} numbers indicate the best results.}
\label{tab:new_vs_ours}
\resizebox{\textwidth}{!}{
\begin{tabular}{l|cccccccccccccc|c}
\toprule
Method & {\sc Ball} & {\sc Bell} & {\sc Bowl} & {\sc Buddha} & {\sc Bunny} & {\sc Cup} & {\sc Die} & {\sc Hippo} & {\sc House} & {\sc Jar} & {\sc Owl} & {\sc Queen} & {\sc Squirrel} & {\sc Tool} & Mean \\
\midrule
Lotus w/ shad. 
& 19.65 & 7.66 & 9.76 & \textbf{14.95} & 9.43 & 20.45 & 12.43 & \textbf{11.00} 
& \textbf{36.54} & 13.06 & 23.16 & \textbf{18.01} & 18.07 & 12.64 & 16.20 \\
Ours
& \textbf{9.09} & \textbf{5.94} & \textbf{6.84} & 17.58 & \textbf{12.70} & \textbf{13.80} & \textbf{8.26} & 14.14
& 36.79 & \textbf{5.93} & \textbf{19.60} & 19.99 & \textbf{21.34} & \textbf{10.66} & \textbf{14.48} \\
\bottomrule
\end{tabular}
}
\end{table}

\blue{\subsection{Comparison with Multi-view Normal Estimation Methods}
We have conducted an additional comparison with multi-view normal estimation methods, including Era3D~\citep{li2024era3d} and Unique3D~\cite{wu2024unique3d}, results are shown in \Tref{tab:mv_compare}. On the LUCES benchmark, Era3D and Unique3D achieve mean angular errors of 43.33° and 23.00$^\circ$, respectively, both substantially worse than the proposed RoSE. Particularly, although Unique3D produces visually high-contrast normal maps, the underlying geometric details are inaccurate, and its performance degrades significantly on objects with complex shadow patterns, which is consistent with our observations of 3D misalignment.}
\begin{table}[ht]
\centering
\caption{Quantitative comparison with multi-view normal estimation methods on the LUCES benchmark. \textbf{Bold} number indicates the best results.}
\label{tab:mv_compare}
\resizebox{\textwidth}{!}{
\begin{tabular}{l|ccccccccccccccc}
\toprule
Method & {\sc Ball} & {\sc Bell} & {\sc Bowl} & {\sc Buddha} & {\sc Bunny} & {\sc Cup} & {\sc Die} & {\sc Hippo} & {\sc House} & {\sc Jar} & {\sc Owl} & {\sc Queen} & {\sc Squirrel} & {\sc Tool} & Mean \\
\midrule
Era3D & 73.02 & 29.53 & 90.85 & 27.85 & 22.82 & 67.50 & 49.73 & 28.32 & 48.87 & 38.92 & 27.86 & 32.24 & 32.13 & 37.03 & 43.33 \\
Unique3D & 14.89 & 13.52 & 13.76 & 24.68 & 16.00 & 34.76 & 19.10 & 16.52 & 44.74 & 15.80 & 28.66 & 26.02 & 26.01 & 27.48 & 23.00 \\
\textbf{Ours} & \textbf{9.09} & \textbf{5.94} & \textbf{6.84} & \textbf{17.58} & \textbf{12.70} & \textbf{13.80} & \textbf{8.26} & \textbf{14.14} & \textbf{36.79} & \textbf{5.93} & \textbf{19.60} & \textbf{19.99} & \textbf{21.34} & \textbf{10.66} & \textbf{14.48} \\
\bottomrule
\end{tabular}
}
\end{table}

\blue{\subsection{Analysis on 3D Reconstruction}
We further evaluate the quality of the reconstructed surfaces obtained from both the predicted and ground truth normal maps on the DiLiGenT and LUCES datasets, using the reconstruction method of~\citep{Cao_2021_CVPR}. The results are reported in~\Tref{tab:recon_diligent} and~\Tref{tab:recon_luces}. By measuring the RMSE~\citep{Cao_2021_CVPR} between the reconstructed and groundtruth surfaces (obtained by reconstruction from the groundtruth normal map), we observe that our method consistently achieves state-of-the-art performance across both benchmarks. These results demonstrate that \ours~not only produces accurate normal maps but also preserves high-fidelity geometric structures after surface reconstruction, further validating the effectiveness of the proposed method.}

\begin{table*}[ht]
\centering
\caption{Quantitative comparison of surface normal RMSE ($\downarrow$) on the DiLiGenT dataset. Results reported with values $\times 10$. \textbf{Bold} number indicates the best results.}
\label{tab:recon_diligent}
\resizebox{\textwidth}{!}{
\begin{tabular}{l|ccccccccccc}
\toprule
Method & {\sc Ball} & {\sc Bear} & {\sc Buddha} & {\sc Cat} & {\sc Cow} & {\sc Goblet} & {\sc Harvest} & {\sc Pot1} & {\sc Pot2} & {\sc Reading} & AVG \\
\midrule
Lotus-G         & 1.08 & 1.31 & 0.55 & 1.19 & 1.58 & 0.84 & 1.83 & 1.96 & 0.84 & 1.18 & 1.24 \\
Lotus-D         & 2.37 & \textbf{0.63} & 0.52 & 1.02 & 1.30 & \textbf{0.76} & 2.25 & 1.91 & 1.01 & \textbf{1.02} & 1.28 \\
Neural LightRig & 1.07 & 0.88 & 1.51 & 2.69 & 1.35 & 1.26 & 5.75 & 2.40 & 0.91 & 1.38 & 1.92 \\
NiRNE           & 1.46 & 0.94 & 0.48 & \textbf{0.77} & 1.47 & 2.12 & 1.54 & \textbf{0.98} & 0.88 & 1.63 & 1.23 \\
\textbf{Ours}   & \textbf{0.79} & 0.88 & \textbf{0.33} & 0.92 & \textbf{0.65} & 1.16 & \textbf{1.24} & 1.35 & \textbf{0.75} & 1.33 & \textbf{0.95} \\
\bottomrule
\end{tabular}
}
\end{table*}

\begin{table}[ht]
\centering
\caption{Quantitative comparison of surface normal RMSE ($\downarrow$) on the LUCES dataset. Results reported with values $\times 10$. \textbf{Bold} number indicates the best results.}
\label{tab:recon_luces}
\resizebox{\textwidth}{!}{
\begin{tabular}{l|ccccccccccccccc}
\toprule
Method & {\sc Ball} & {\sc Bell} & {\sc Bowl} & {\sc Buddha} & {\sc Bunny} & {\sc Cup} & {\sc Die} & {\sc Hippo} & {\sc House} & {\sc Jar} & {\sc Owl} & {\sc Queen} & {\sc Squirrel} & {\sc Tool} & AVG \\
\midrule
Lotus-G         & 0.89 & 0.36 & 1.33 & 1.12 & 0.72 & 1.18 & 0.67 & 1.23 & 1.23 & 0.57 & \textbf{0.83} & 0.55 & 1.30 & 0.34 & 0.88 \\
Lotus-D         & 0.89 & 0.38 & 1.09 & 1.13 & 0.73 & 1.82 & 0.75 & 1.09 & 1.54 & 0.56 & 0.89 & 0.46 & 0.86 & 0.31 & 0.89 \\
Neural LightRig & \textbf{0.31} & 0.51 & 3.02 & \textbf{0.57} & 0.69 & 0.98 & 1.71 & 1.51 & 1.28 & 1.88 & 1.05 & 0.45 & 0.60 & 0.65 & 1.09 \\
NiRNE           & 0.77 & 0.76 & 2.12 & 0.70 & 0.64 & 1.10 & 1.02 & 1.71 & 1.12 & 0.75 & 1.93 & 0.75 & 0.71 & 0.71 & 1.05 \\
\textbf{Ours}   & 0.34 & \textbf{0.19} & \textbf{0.80} & 0.69 & \textbf{0.53} & \textbf{0.72} & \textbf{0.54} & \textbf{0.79} & \textbf{0.67} & \textbf{0.42} & 1.22 & \textbf{0.25} & \textbf{0.54} & \textbf{0.22} & \textbf{0.57} \\
\bottomrule
\end{tabular}
}
\end{table}

\blue{\subsection{Performance on Normal Estimation using Video Diffusion Model}
We have conducted an additional comparison with the variant that using SV3D to directly predict the single-frame surface normal (noted as ``SVD-nml''), the results are shown in \Tref{tab:svd-nml}. The average result on LUCES benchmark dataset is $20.61^\circ$, indicating that the dense information geometrically encoded in the video model does not play a critical role in this setting, and the model also loses its ability to make use of temporal information.}

\begin{table}[ht]
\centering
\caption{Quantitative comparison with ``SVD-nml'' on the LUCES benchmark. \textbf{Bold} number indicates the best results.}
\label{tab:svd-nml}
\resizebox{\textwidth}{!}{
\begin{tabular}{l|ccccccccccccccc}
\toprule
Method & {\sc Ball} & {\sc Bell} & {\sc Bowl} & {\sc Buddha} & {\sc Bunny} & {\sc Cup} & {\sc Die} & {\sc Hippo} & {\sc House} & {\sc Jar} & {\sc Owl} & {\sc Queen} & {\sc Squirrel} & {\sc Tool} & Mean \\
\midrule
SVD-nml & 17.20 & 9.96 & 19.69 & 22.28 & 14.77 & 18.32 & 11.47 & 18.23 & 49.74 & 10.49 & 26.95 & 25.04 & 26.03 & 18.29 & 20.61 \\
\textbf{Ours} & \textbf{9.09} & \textbf{5.94} & \textbf{6.84} & \textbf{17.58} & \textbf{12.70} & \textbf{13.80} & \textbf{8.26} & \textbf{14.14} & \textbf{36.79} & \textbf{5.93} & \textbf{19.60} & \textbf{19.99} & \textbf{21.34} & \textbf{10.66} & \textbf{14.48} \\
\bottomrule
\end{tabular}
}
\end{table}